\documentclass[runningheads]{llncs}

\usepackage{eccv}

\usepackage{eccvabbrv}

\usepackage{graphicx}
\usepackage{booktabs}

\usepackage{amsmath}
\usepackage{caption}
\usepackage{multirow}
\usepackage{makecell}
\usepackage{subcaption}
\usepackage{amssymb}
\usepackage{wrapfig}

\usepackage{hyperref}

\usepackage{orcidlink}

\begin{document}

\title{Intrinsically Stable Spiking Neural Networks: Overcoming the Performance Barrier in the Absence of Batch Normalization}

\titlerunning{IS-SNN}

\author{Ruichen Ma\inst{1}\orcid{0009-0004-6210-8022} \and
Xiaoyang Zhang\inst{2}\orcid{0009-0005-8084-2906} \and
Jian Bai\inst{2}\orcid{0000-0002-9845-6991} \and
Guanchao Qiao\inst{1}\orcid{0000-0003-4982-5938} \and
Liwei Meng\inst{1}\orcid{0009-0002-0842-1224} \and
Ning Ning\inst{1} \and
Yang Liu\inst{1} \and
Shaogang Hu\inst{1,3}\thanks{Corresponding author. (Email: sghu@uestc.edu.cn)}\orcid{0000-0002-8653-2491}}

\authorrunning{Ma et al.}

\institute{University of Electronic Science and Technology of China (UESTC) \and
Beijing Institute of Remote-Sensing Equipment \and
Shenzhen Institute for Advanced Study, UESTC \\
\url{https://github.com/Ruichen0424/IS-SNN}
}

\maketitle

\begin{abstract}

The performance of deep spiking neural networks (SNNs) often relies on batch normalization (BN).
However, the advanced dynamic BN variants used in state-of-the-art models introduce runtime multiplications, which weaken the hardware-efficiency motivation of SNNs.
To address this tension, we identify catastrophic firing-rate decay as a primary cause of severe performance degradation in normalization-free SNNs.
Guided by this insight, this work proposes the Intrinsically Stable SNN (IS-SNN) architecture, which removes activation-normalization layers by enforcing signal homeostasis through topology-aware weight standardization and modified residual connections.
By folding the standardization operations into static weights offline, IS-SNN removes the runtime statistics tracking and multiplications introduced by activation normalization, restoring an accumulation-oriented inference datapath.
Comprehensive experiments show that IS-SNN achieves performance competitive with or superior to computationally expensive dynamic BN techniques across VGG, ResNet, and Transformer-based models.
Notably, it achieves a competitive accuracy of 68.05\% on ImageNet and overcomes the severe depth limitations of prior BN-free attempts.
Together with a 96.4\% reduction in FPGA lookup table resource consumption for neuron implementations, these results support IS-SNN as a practical framework for building accurate and hardware-friendly deep neuromorphic systems.

\end{abstract}

\section{Introduction}
\label{sec:Introduction}

Spiking neural networks (SNNs), inspired by the information processing mechanisms of the brain, operate using asynchronous, event-driven spikes \cite{xu2018csnn, zenke2021visualizing, ma2026i2e}.
This paradigm offers the potential to realize a new generation of highly energy-efficient intelligent systems \cite{roy2019towards}.
When implemented on neuromorphic hardware, SNNs can achieve substantially lower power consumption than conventional artificial neural networks (ANNs) \cite{merolla2014million, furber2014spinnaker, qiao2015reconfigurable, davies2018loihi, imam2020rapid}.
While early approaches successfully converted pre-trained ANNs to SNNs \cite{hao2023bridging, jiang2023unified, wang2023new, hao2023reducing}, they typically suffered from high inference latency.
Consequently, direct training via surrogate gradients has emerged as a dominant methodology for low-latency inference \cite{wu2018spatio, neftci2019surrogate}, delivering highly competitive accuracy.

However, the success of state-of-the-art deep SNNs has become increasingly dependent on specialized batch normalization (BN) variants to stabilize training and improve performance \cite{sengupta2019going, wu2019direct, zheng2021going, kim2021revisiting, deng2022temporal, duan2022temporal, ikegawa2022rethinking, jiangtab}.
This reliance creates a tension with the hardware-efficiency motivation of neuromorphic computing.
While standard BN in SNNs can be fused into preceding synaptic weights or neuronal thresholds during inference \cite{guo2023membrane, qiao2023batch}, leading models often rely on advanced, batch- or time-dependent BN variants such as TEBN \cite{duan2022temporal} and TAB \cite{jiangtab}.
These dynamic methods are difficult to fold into static parameters because their normalization statistics must be recomputed at runtime.
Although advanced neuromorphic chips technically support multiplication operations, these incur substantially higher area and energy costs than simple adders \cite{chen2021bnn, ma2024b}.
By reintroducing runtime multiplications and statistical tracking at every timestep, dynamic BN weakens the accumulation-oriented datapath that motivates SNN deployment.

This dependency creates an important dilemma: simply removing activation normalization is not a viable solution.
As corroborated by forward signal analysis, deep SNNs lacking statistics control face a signal-propagation barrier: catastrophic firing-rate decay.
Without normalization, pre-activation variances can drift across layers, causing neuronal firing rates to either vanish or saturate.
This disrupts signal propagation and can lead to severe performance degradation or training collapse.
Consequently, the field faces an undesirable trade-off between using costly, non-fusible dynamic normalization and accepting severe limitations on network depth.
Prior attempts at BN-free SNNs have been largely restricted to shallow models and simpler datasets and have not yet established an effective solution for modern deep architectures \cite{qiao2023batch}.

To address this conflict, we shift the perspective from optimization instability to signal survival.
Motivated by the observation that catastrophic firing-rate decay is a primary bottleneck in normalization-free SNNs, this paper introduces the Intrinsically Stable SNN (IS-SNN).
IS-SNN removes activation-normalization layers while maintaining competitive accuracy through intrinsic signal stabilization.
Building on weight reparameterization and residual scaling, IS-SNN adapts these ideas to spiking dynamics through topology-aware weight standardization and modified residual connections that enforce signal homeostasis.
By decoupling the training and inference phases, IS-SNN enables the standardization operations to be folded into static weights offline before deployment.
The main contributions of this work are as follows:
\begin{itemize}
    \item The proposed IS-SNN maintains signal homeostasis with topology-aware weight standardization and a modified residual connection. By decoupling training and inference, the standardization operations can be folded into static weights offline, introducing no inference-time normalization overhead.
    \item Comprehensive experiments demonstrate that IS-SNN achieves performance competitive with or superior to advanced, computationally expensive dynamic BN variants across diverse architectures, including VGG, ResNet, and Transformer-based models. Notably, it scales to deep networks such as ResNet-152 and achieves 68.05\% accuracy on ImageNet, substantially improving over naive BN-free baselines.
    \item A dedicated hardware analysis shows a 15\% increase in training throughput and a 96.4\% reduction in FPGA lookup table (LUT) resource consumption for neuron implementations, supporting the practical efficiency benefits of removing runtime normalization operations.
\end{itemize}


\section{Related Work}
\label{sec:Related_Work}

\paragraph{Normalization in SNNs and Hardware Fusibility.}
Batch normalization (BN) \cite{ioffe2015batch} has been central to training high-performance neural networks.
In SNNs, early adaptations included standard BN \cite{fang2021deep} and temporal extensions \cite{shrestha2018slayer}.
Standard static BN can be mathematically folded into synaptic weights or neuronal thresholds during inference, thereby preserving the inference efficiency of SNNs \cite{guo2023membrane, qiao2023batch}.
However, to close the accuracy gap with ANNs, modern deep SNNs often rely on advanced time- or batch-dependent BN variants, such as tdBN \cite{zheng2021going}, BNTT \cite{kim2021revisiting}, TEBN \cite{duan2022temporal}, and TAB \cite{jiangtab}.
Because these methods compute normalization statistics dynamically at runtime, they are difficult to reduce to fixed offline parameters.
This reintroduces runtime multiplications and weakens the accumulation-oriented hardware efficiency that motivates SNN deployment.

\paragraph{Activation-Normalization-Free SNNs.}
The overhead of dynamic BN has motivated research into BN-free SNNs.
OTTT \cite{xiao2022online} uses Weight Standardization (WS), following normalization-free ANN designs, in the context of online training through time.
Its primary goal is to reduce online training complexity, whereas our work addresses the different problem of stabilizing deep SNNs after removing activation normalization.
Rather than treating WS as a generic weight reparameterization, IS-SNN derives topology-aware scaling factors from SNN signal propagation, neuron output variance, and residual variance growth under spiking dynamics.
Other BN-free SNN attempts \cite{qiao2023batch} have mainly been evaluated on shallow VGG architectures and small-scale datasets.
Establishing a general framework for high-performance, activation-normalization-free SNNs that scales to modern deep architectures therefore remains an important open challenge.

\paragraph{Weight Reparameterization.}
Weight reparameterization provides another route to improving training stability without activation normalization.
In conventional ANNs, Weight Standardization \cite{qiao2019micro} was first introduced to smooth the loss landscape and accelerate optimization, particularly for micro-batch training.
Normalizer-free ResNet studies further developed signal propagation analysis, scaled weight standardization, and residual branch scaling for deep ANNs without BN \cite{brock2021characterizing, brock2021high}.
These works provide important foundations for normalization-free network design.
However, directly transferring these principles to SNNs is non-trivial because spike activations are discrete, temporally integrated, and governed by neuron-specific firing dynamics.
In IS-SNN, WS is therefore adapted to maintain firing-rate homeostasis.
The scaling coefficient $\gamma_\ell$ is derived from SNN-specific variance propagation, including the empirically estimated neuron output variance $\sigma_g$ and topology-dependent residual accumulation.
As confirmed by our ablations, naive WS is insufficient for deep SNNs, whereas this topology-aware formulation effectively counteracts firing-rate decay while allowing the standardization operations to be folded into static weights before deployment.

\section{Method}
\label{sec:Method}
\subsection{Preliminaries of Spiking Neuron Dynamics}
The core computational unit of an SNN is the spiking neuron (SN).
Its dynamics encompass three stages: integration, firing, and reset:
\begin{gather}
H[t]=f(V[t-1], X[t]) \label{equ_dynamics} \\
S[t]=\Theta(H[t]-V_{th}) \\
V[t]=H[t](1-S[t])+V_{reset}S[t]
\end{gather}
where $X[t]$ is the input current at timestep $t$.
$H[t]$ and $V[t]$ denote the membrane potentials immediately before and after spike generation, respectively.
The output spike $S[t]$ is a binary event triggered when $H[t]$ exceeds the firing threshold $V_{th}$.
The Heaviside step function is defined as $\Theta(x)=1$ for $x>0$ and $\Theta(x)=0$ for $x\leq0$.
Following a spike, the membrane potential resets to $V_{reset}$.

The integration function $f(\cdot)$ in \cref{equ_dynamics} defines the specific neuronal behavior.
The standard Leaky Integrate-and-Fire (LIF) neuron is formulated as:
\begin{equation}
H[t]_{LIF} = V[t-1] - \frac{1}{\tau}(V[t-1] - V_{reset}) + X[t]
\label{no_decay_input}
\end{equation}
where $\tau$ is the membrane time constant.
For neuromorphic datasets, the Parametric LIF (PLIF) model \cite{fang2021incorporating} was employed, where $\tau$ is learnable.
The LIF model with decaying input similarly applies a $1/\tau$ coefficient to $X[t]$.

\subsection{Forward Signal Analysis of Firing Rate}
\label{sec:3.2}

To investigate the failure mechanisms of deep SNNs trained without activation normalization, forward signal analysis was performed to track activation statistics during propagation.
This analysis is important for networks that lack active mechanisms to regulate internal data distributions.
In SNNs, the mean activation $\mu$ corresponds directly to the average firing rate, physically representing the network's overall activity level.

A deep neural network can be conceptualized as a sequence of stacked minimum repetition units (MRUs).
In plain networks like VGG, an MRU consists of a Conv-BN-SN sequence.
In residual networks, it comprises several residual blocks and a transition block.
For information to propagate effectively, the firing rate should remain stable across successive MRUs.
If the firing-rate ratio between adjacent MRUs, $c = \mu_{L+1}/\mu_L$, deviates substantially from $1$, the rate will either vanish or saturate as the layer depth $L$ increases:
$\lim_{L \to +\infty}\mu_L=0$ (when $0\leq c<1$) or $\lim_{L \to +\infty}\mu_L=1$ (when $c>1$).
Both outcomes are detrimental.
A vanishing rate indicates a broken signal path, while a saturated rate limits the network's expressive capacity.
From an information-theoretic perspective, the entropy of the spike train, $H(S_L)=-\mu_L\log\mu_L-(1-\mu_L)\log(1-\mu_L)$, is maximized when $\mu_L=0.5$ and approaches zero as $\mu_L$ approaches either $0$ or $1$.

\begin{figure}[!t]
  \centering
  \begin{subfigure}{0.2528\linewidth}
    \centering
    \includegraphics[width=\linewidth]{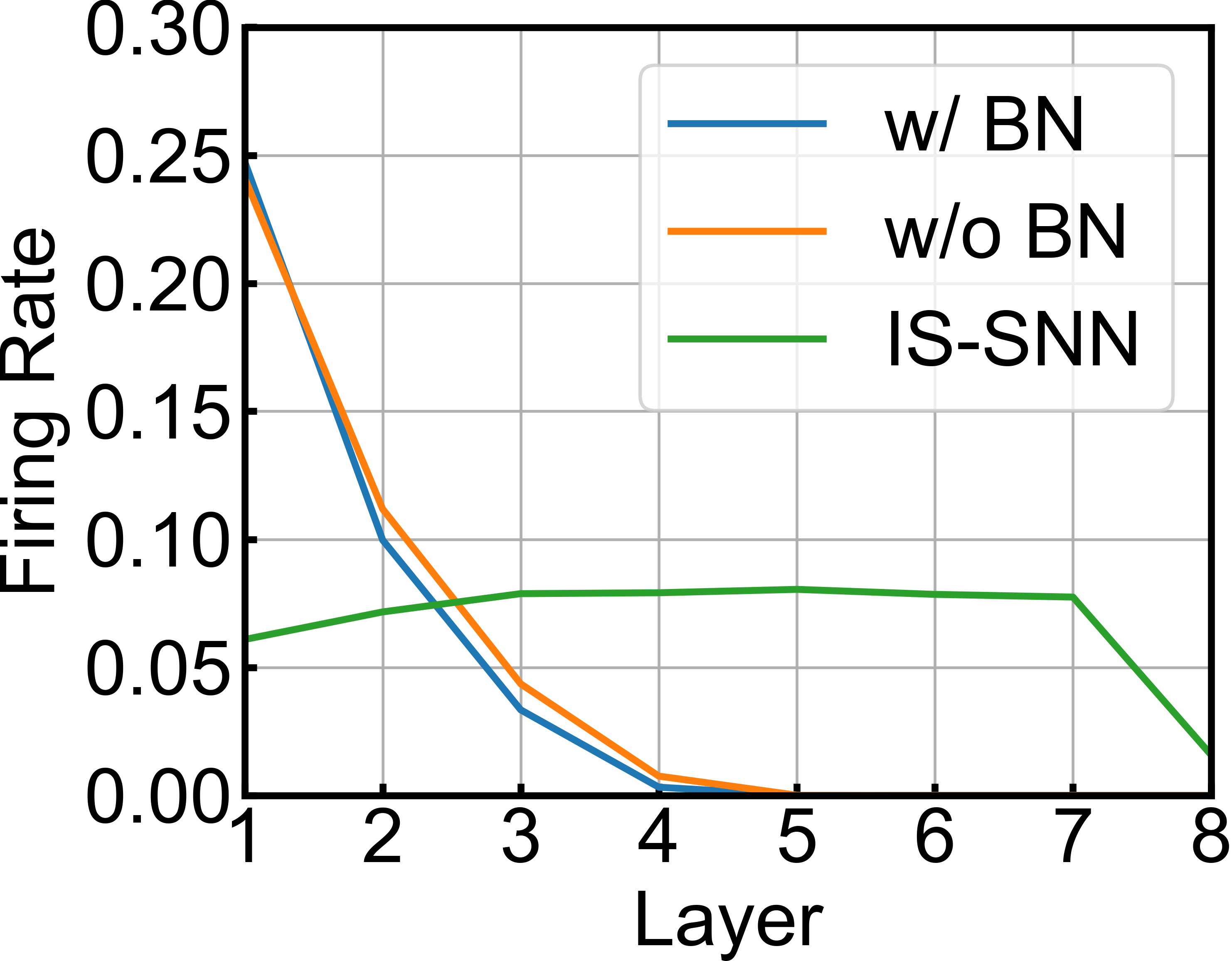}
    \caption{Before training.}
    \label{fig:VGG9_FR_a}
  \end{subfigure}
  \begin{subfigure}{0.2355\linewidth}
    \centering
    \includegraphics[width=\linewidth]{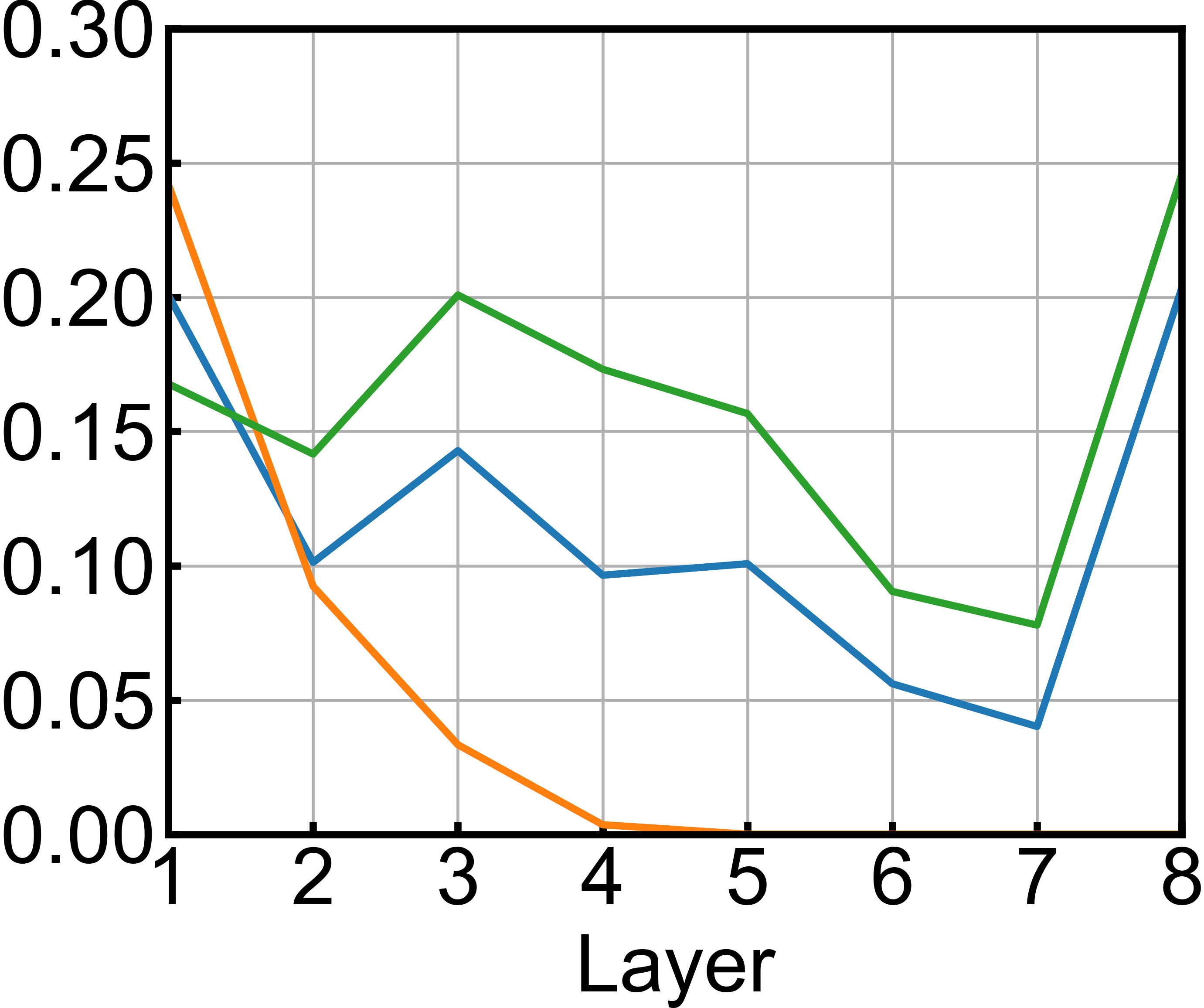}
    \caption{After training.}
    \label{fig:VGG9_FR_b}
  \end{subfigure}
  \hfill
  \begin{subfigure}{0.2572\linewidth}
    \centering
    \includegraphics[width=\linewidth]{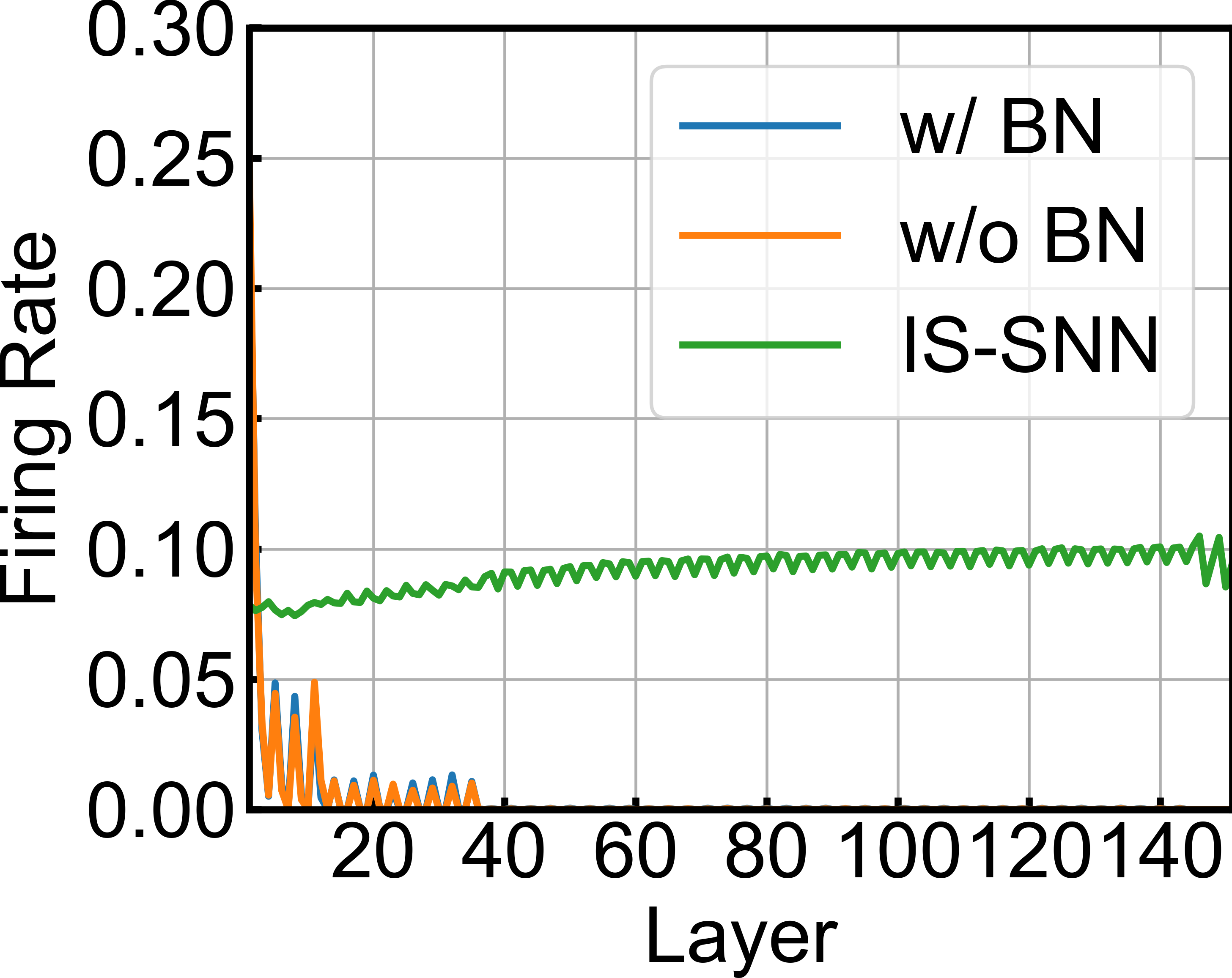}
    \caption{Before training.}
    \label{fig:SEW152_FR_a}
  \end{subfigure}
  \begin{subfigure}{0.2302\linewidth}
    \centering
    \includegraphics[width=\linewidth]{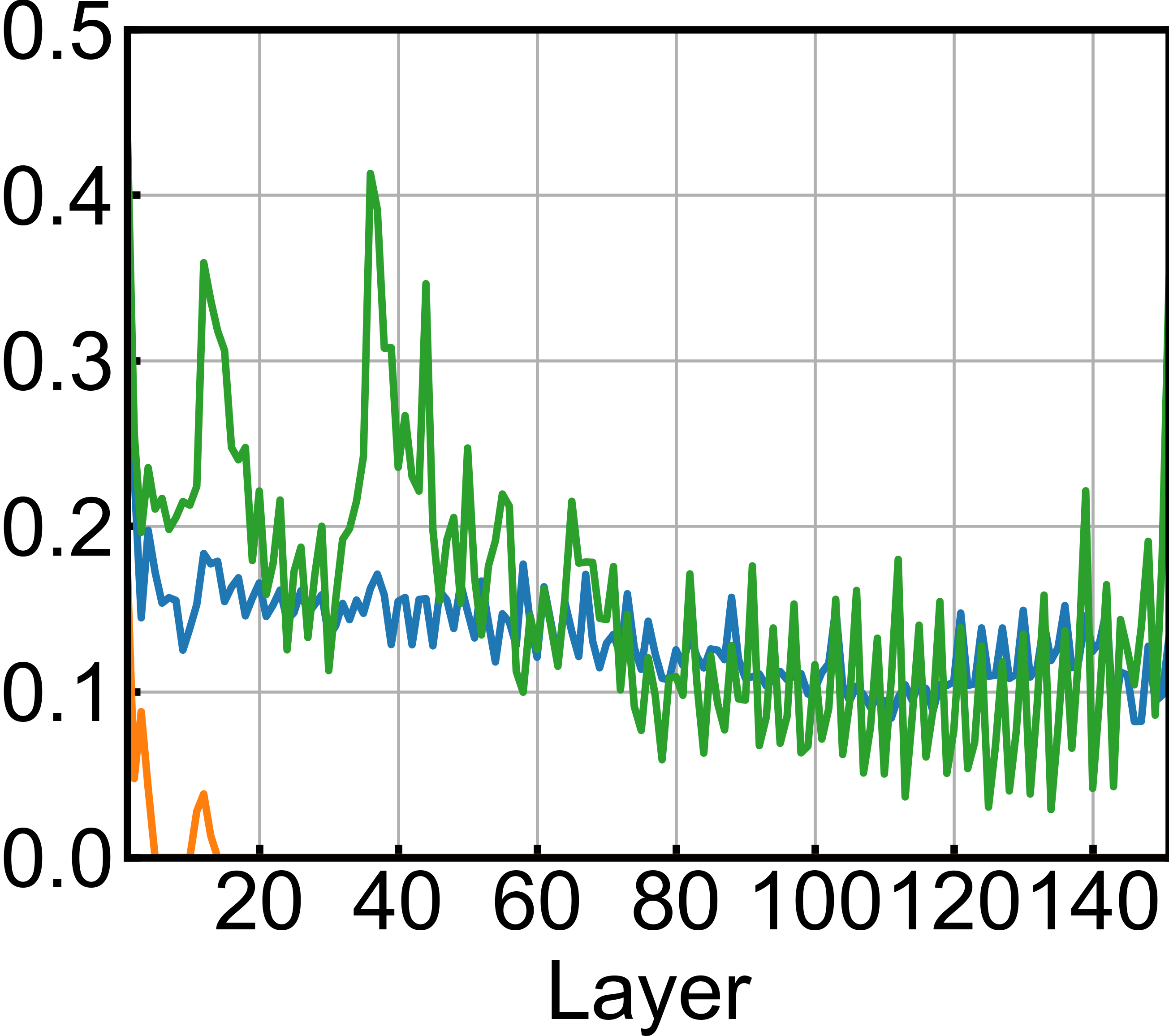}
    \caption{After training.}
    \label{fig:SEW152_FR_b}
  \end{subfigure}
  \caption{Layer-wise firing rates in SNNs. (a-b) VGG-9 on CIFAR-10 before and after training. Without BN, the network suffers from rapid firing-rate decay and fails to recover. Both the BN baseline and the proposed IS-SNN prevent this collapse. (c-d) SEW-ResNet-152 on CIFAR-100. A similar signal-decay phenomenon is observed in the deep architecture, where IS-SNN maintains stable signal propagation without BN.}
  \label{fig:firing_rates_combined}
\end{figure}

This failure mode is empirically supported by \cref{fig:firing_rates_combined}.
The figures plot the layer-wise firing rates for VGG-9 on CIFAR-10 and SEW-ResNet-152 on CIFAR-100, comparing models with BN, without BN, and with the proposed IS-SNN.
Without BN, the naive baseline suffers from firing-rate decay across layers and fails to recover.
In contrast, BN-equipped networks dynamically restore rates to a stable range during training.
These results indicate that catastrophic firing-rate decay is a primary cause of signal collapse in deep SNNs lacking activation normalization.
The key role of BN is to provide a mechanism for dynamic stabilization.
This is analogous to homeostatic plasticity in biological systems, where neurons dynamically regulate their properties to maintain firing rates within a stable range \cite{turrigiano1999homeostatic}.
This parallel suggests that maintaining a stable firing rate is also important for efficient information transmission in artificial spiking systems.

\subsection{The IS-SNN Architecture}
\label{sec:unnormalized_structure}
Motivated by the preceding analysis, the IS-SNN architecture is designed to stabilize the firing rate by controlling the pre-activation statistics layer by layer.
Inspired by synaptic scaling, a biological homeostatic process that adjusts synaptic strengths to stabilize neuronal activity \cite{turrigiano2004homeostatic}, IS-SNN employs Weight Standardization (WS) to remove data-dependent activation-normalization layers.
This formulation decouples the training and inference phases, enabling deployment without inference-time normalization overhead.

\subsubsection{Weight Standardization and Offline Reparameterization.}
To mimic the stabilizing effect of BN, IS-SNN aims to keep the pre-activation scale close to a standard-normal reference, $x_{pa}\sim\mathcal{N}(0, 1)$.
The mean $\mu_{pa}$ and variance $\sigma_{pa}^2$ of the pre-activation are governed by the statistics of input spikes and weights:
\begin{gather}
    \mu_{pa}=N\mu_{in}\mu_{W_i} \label{equ_miu} \\
    \sigma_{pa}^2=N\sigma_{in}^2(\sigma_{W_i}^2+\mu_{W_i}^2) \label{equ_sigma}
\end{gather}
where $N$ is the fan-in.
To approach the target scale $x_{pa}\sim\mathcal{N}(0, 1)$, the weights are regulated to maintain a zero mean, $\mu_{W_i} = 0$, and a scaled variance, $\sigma_{W_i}^2 = 1/(N\sigma_{in}^2)$.
We enforce these conditions using weight standardization:
\begin{equation}
    \hat{W}_{i,j}=\gamma_{\ell}\cdot\frac{W_{i,j}-\mu_i}{\sqrt{N\sigma_i^2+\epsilon}}
\label{eq:train_ws}
\end{equation}
where $W_{i,j}$ are the original weights, $\mu_i$ and $\sigma_i^2$ are their statistical mean and variance, $\hat{W}_{i,j}$ are the standardized weights, and $\epsilon=10^{-4}$ ensures numerical stability.
The critical component of this formulation is the scaling factor $\gamma_{\ell}$, which must be set appropriately for each layer $\ell$ to control the pre-activation variance.
A convolutional layer incorporating this technique is termed WS-Conv.

During training, the learnable weights $W_{i,j}$ continuously update, causing their statistics $\mu_i$ and $\sigma_i^2$ to fluctuate.
Therefore, at the beginning of each training step, the original weights are dynamically standardized using \cref{eq:train_ws}.
The network then uses the standardized weights $\hat{W}_{i,j}$ for forward propagation.
Once training concludes, all parameters, including the weights, their statistics, and the scaling factor $\gamma_\ell$, become fixed.
This enables the standardization operations to be folded into static weights offline before deployment.
During inference, the hardware uses these pre-scaled static weights $\hat{W}_{i,j}$.
By decoupling normalization from the inference phase, the deployed IS-SNN is structurally identical to a raw BN-free network.
It requires no runtime normalization statistics tracking or normalization-related multiplications, preserving the accumulation-oriented datapath of SNNs.

\begin{figure}[!t]
  \centering
  \begin{minipage}[]{0.48\linewidth}
    \centering
    \includegraphics[width=0.8\linewidth]{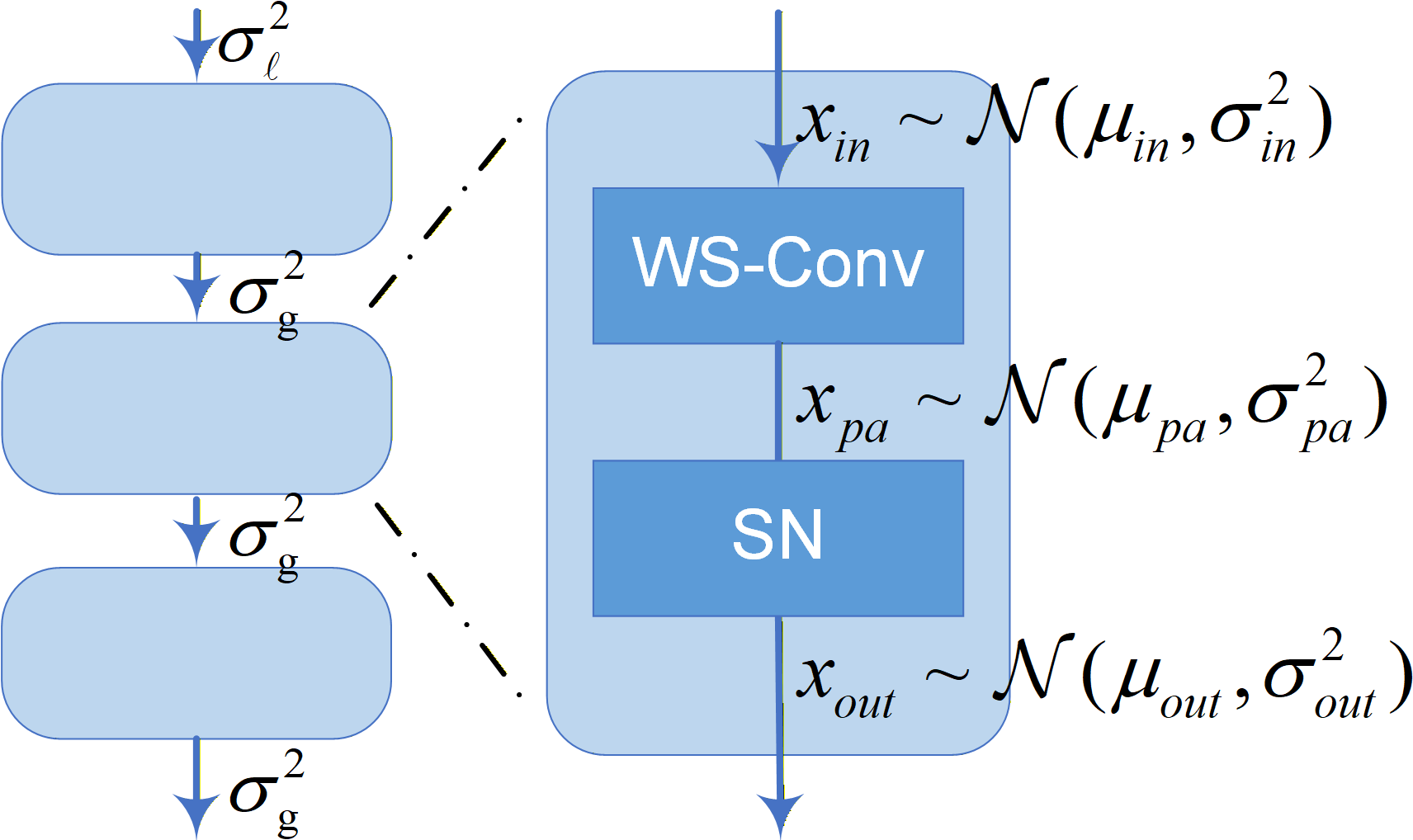}
    \caption{The minimum repetition unit (MRU) of the plain IS-SNN, which replaces the Conv-BN-SN block with a weight-standardized convolutional layer followed by a spiking neuron.}
    \label{VGG_MRU}
  \end{minipage}
  \hfill
  \begin{minipage}[]{0.49\linewidth}
    \centering
    \begin{subfigure}{0.23\linewidth}
      \centering
      \includegraphics[width=0.95\linewidth]{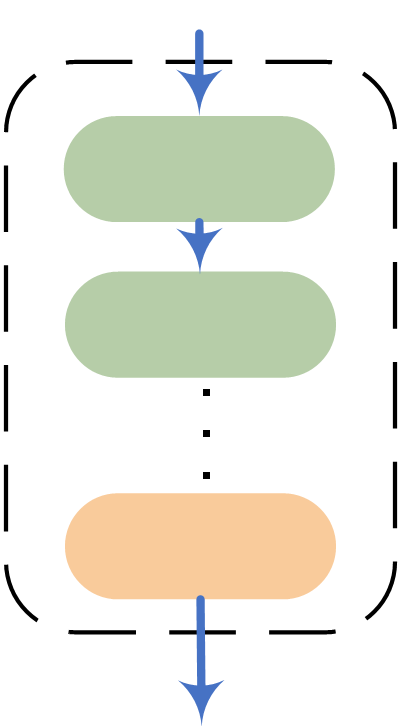}
	  \caption{}
      \label{SEW_MRU_3}
    \end{subfigure}
    \begin{subfigure}{0.375\linewidth}
      \centering
      \includegraphics[width=\linewidth]{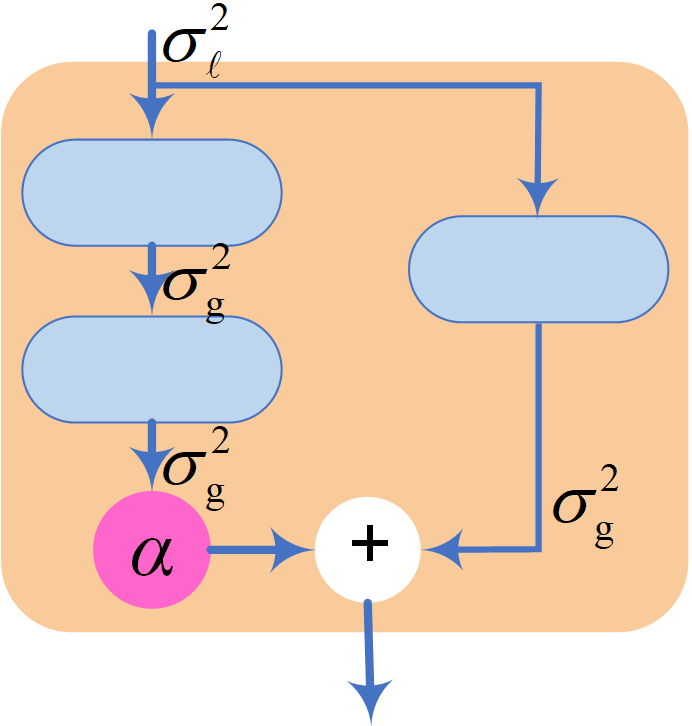}
	  \caption{}
      \label{SEW_MRU_1}
    \end{subfigure}
    \begin{subfigure}{0.355\linewidth}
      \centering
      \includegraphics[width=0.89\linewidth]{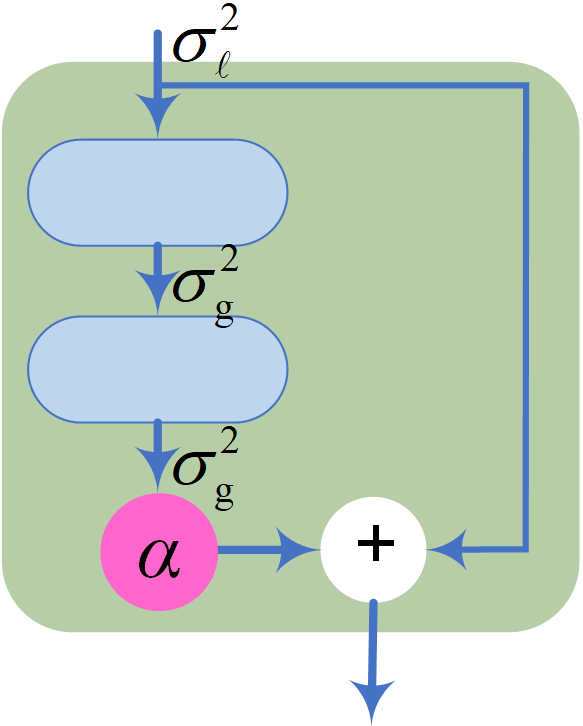}
	  \caption{}
      \label{SEW_MRU_2}
    \end{subfigure}
    \caption{The (a) MRU of the residual IS-SNN consists of one (b) transition block and several (c) residual blocks. The modified residual connection incorporates a scaling factor $\alpha$ to control variance growth.}
    \label{SEW_MRU}
  \end{minipage}
\end{figure}

\subsubsection{Topology-Aware Variance Propagation.}
With the WS mechanism established, its effectiveness depends on determining the scaling factor $\gamma_{\ell}$ for each layer across different network topologies.
For clarity in derivation, we introduce three core parameters governing variance propagation:
\begin{itemize}
\item $\sigma_\ell$: The theoretically estimated input standard deviation for layer $\ell$. It is a topology-dependent constant and remains invariant across timesteps.
\item $\gamma_\ell$: The layer-wise scaling factor, defined as $\gamma_\ell \equiv 1 / \sigma_\ell$, which ensures the weights satisfy the variance condition $\sigma_{W_i}^2 = 1/(N\sigma_{\ell}^2)$ required by \cref{eq:train_ws}.
\item $\sigma_g$: A fixed constant representing the empirically estimated output standard deviation of a specific spiking neuron type under a standard-normal input. It is an intrinsic neuronal property, invariant across layers and timesteps.
\end{itemize}

The calculation of $\sigma_\ell$ depends on the network architecture.
In plain non-residual networks such as VGG, the output of layer $\ell$ serves directly as the input to layer $\ell+1$, as shown in \cref{VGG_MRU}.
Consequently, the variance remains constant across layers: $\sigma_{in, \ell+1}^2 = \sigma_{out, \ell}^2 = \sigma_g^2$.
The scaling factor for any layer $\ell$ is therefore set to $\gamma_{\ell}\equiv1/\sigma_g$.
Conversely, residual networks compound variance through addition operations.
This can cause signal variance to grow with successive blocks and destabilize BN-free training.
To control this growth, we introduce a modified residual connection, as shown in \cref{SEW_MRU}:
\begin{equation}
x_{\ell+1} = \alpha\cdot\text{SN}(f(x_{\ell})) + x_{\ell}
\end{equation}
where $\alpha$ is a scaling coefficient that regulates the residual branch's contribution (uniformly set to $0.5$ across all architectures, as justified in \cref{sec:4.3}).
Within a continuous sequence of residual blocks, the tracked input variance follows $\sigma_{\ell+1}^2 = \alpha^2\cdot\sigma_g^2 + \sigma_{\ell}^2$.
At the beginning of a stage, the transition block resets the statistics to $\sigma_{reset}^2=(1+\alpha^2)\cdot\sigma_g^2$.
By analytically tracking the propagated variance $\sigma_{\ell}^2$ layer by layer, the scaling factor $\gamma_{\ell}=1/\sigma_{\ell}$ can be computed throughout the ResNet topology.


\subsubsection{Empirical Estimation of $\sigma_g^2$.}
The design of IS-SNN depends on the neuron's base output variance $\sigma_g^2$.
Unlike ReLU units, which have a transparent linear relationship between input and output variance \cite{brock2021characterizing}, the dynamics of spiking neurons are highly non-linear and timestep-dependent, making analytical derivation difficult.
Therefore, $\sigma_g^2$ is approximated empirically by stimulating various neuron models with random pre-activations drawn from $\mathcal{N}(0,1)$.
\Cref{table_sigma} lists the time-averaged output variance $\overline{\sigma_g^2}$ for different LIF neuron configurations.
This simulation-based method can be applied to estimate the corresponding $\sigma_g^2$ for other neuron models.
See the Appendix for more details.

\subsubsection{Impact of Auxiliary Network Components.}
\begin{figure}[t]
  \centering
  \begin{minipage}[]{0.49\linewidth}
    \centering
    \captionof{table}{Empirically estimated time-averaged output variance, $\overline{\sigma_g^2}$, for different Leaky Integrate-and-Fire (LIF) models.}
    \label{table_sigma}
    \begin{tabular}{p{3.1em} p{3.2em}<{\centering} p{3.1em}<{\centering} p{3.1em}<{\centering} p{3.1em}<{\centering}}
        \toprule[1.5pt]
        \multirow{2.5}{*}{Model} & \multirow{2.5}{*}{\makecell{Decay \\ Input}} & \multicolumn{3}{c}{$\overline{\sigma_{g}^2}$} \\
        \cmidrule{3-5}
         & & T=4 & T=8 & T=16 \\
        \midrule[1pt]
        \multirow{2}{*}{\makecell{LIF\\($\tau=2$)}} & $\times$ & 0.1234 & 0.1195 & 0.1174 \\
         & $\surd$ & 0.0290 & 0.0298 & 0.0302 \\
        \bottomrule[1.5pt]
    \end{tabular}
  \end{minipage}
  \hfill
  \begin{minipage}[]{0.49\linewidth}
    \centering
    \begin{subfigure}{0.524\linewidth}
      \includegraphics[width=\linewidth]{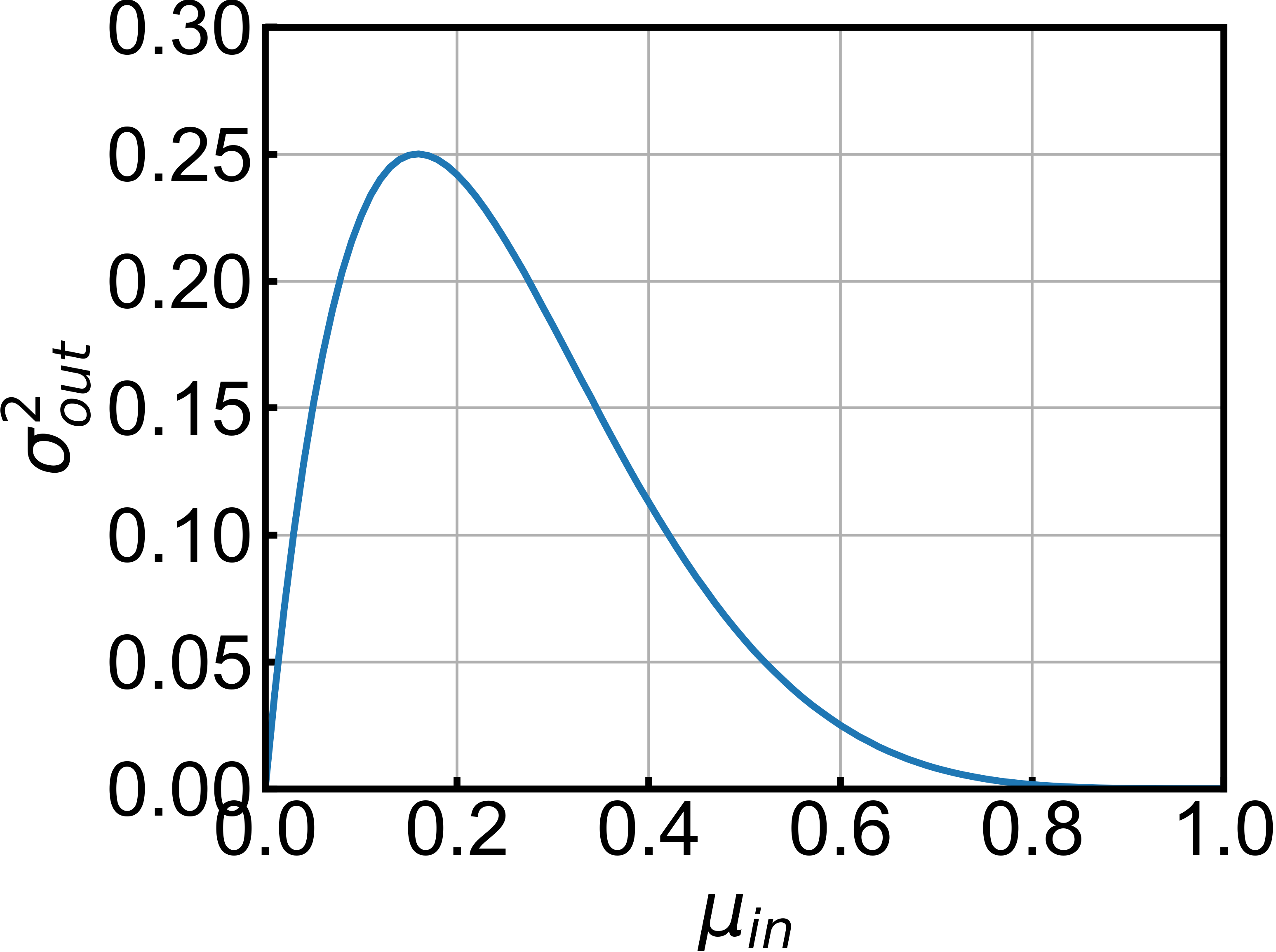}
      \caption{$\sigma_{out}^2-\mu_{in}$}
      \label{maxpool1}
    \end{subfigure}
    \begin{subfigure}{0.43\linewidth}
      \includegraphics[width=\linewidth]{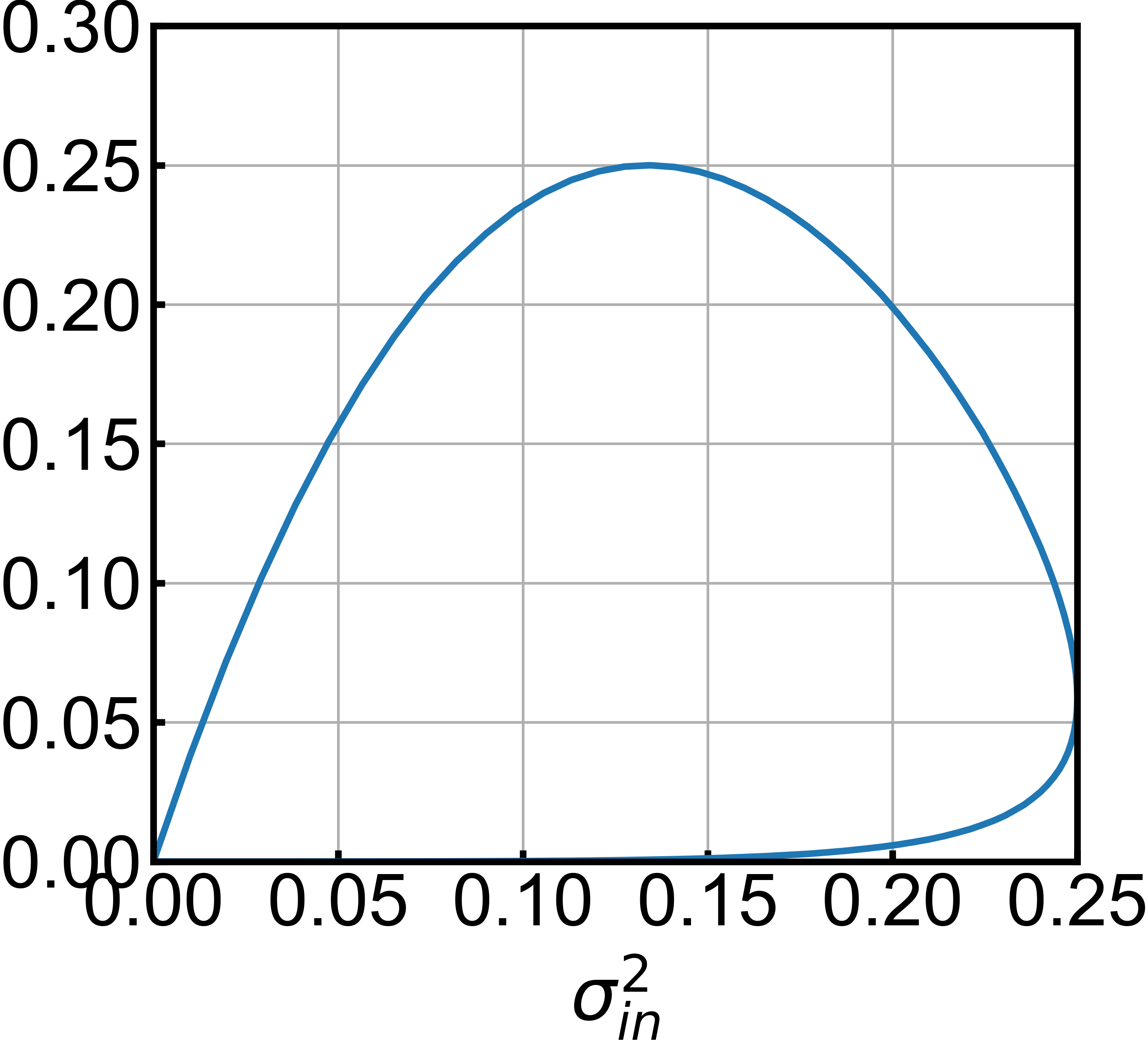}
      \caption{$\sigma_{out}^2-\sigma_{in}^2$}
      \label{maxpool2}
    \end{subfigure}
    \caption{Simulated output variance $\sigma_{out}^2$ of a max-pooling layer processing binary spike inputs.}
    \label{maxpool}
  \end{minipage}
\end{figure}

Signal propagation is also affected by other network components, such as max-pooling layers.
The relationship between input and output variance is particularly complex for binary spike activations.
As the input firing rate $\mu_{in}$ increases from 0 to 1, the input variance $\sigma_{in}^2$ first increases from 0 to a maximum of 0.25 before decreasing back to 0.
This complex, non-monotonic relationship is detailed in \cref{maxpool}.
While max-pooling is supported in IS-SNN, caution is recommended due to this complex interaction.
Downsampling is more reliably achieved using strided convolutions.
For the first encoding layer, the scaling factor $\gamma_1$ is uniformly set to 1 for simplicity and generalizability.
By analyzing the effect of each component, the theoretically expected input variance $\sigma_{\ell}^2$ for each layer can be estimated, completing the specification of the IS-SNN architecture.

\section{Experiments}
\label{sec:Experiments}


\begin{table}[!t]
	\caption{Performance comparison of IS-SNN against state-of-the-art normalization methods and a baseline without normalization (w/o BN). The w/o BN results show severe performance degradation on deep or complex models. IS-SNN achieves accuracy competitive with or superior to advanced, non-fusible normalization methods while introducing no inference-time normalization overhead.}
	\label{table_com1}
  \renewcommand{\arraystretch}{0.55}
	\centering
	\begin{tabular}{p{6.7em} p{7.7em}<{\centering} p{5.0em}<{\centering} p{5.0em}<{\centering} p{11.5em}<{\centering}}
		\toprule[1.5pt]
		{Dataset} & {Model} & {Method} & {Timesteps} & {Accuracy (\%)} \\
		\midrule[1pt]
		\multirow{11}{*}{CIFAR-10} & Spike-Norm \cite{sengupta2019going} & VGG-16 & 2500 & 91.55 \\
		 & tdBN \cite{zheng2021going} & ResNet-19 & 6 / 4 / 2 & 93.16 / 92.92 / 92.34 \\
		 & BNTT \cite{kim2021revisiting} & VGG-9 & 25 / 20 & 90.5 / 90.3 \\
		 & TET \cite{deng2022temporal} & ResNet-19 & 6 / 4 / 2 & 94.50 / 94.44 / 94.16 \\
		\cmidrule{2-5}
		 & {\multirow{2}{*}{w/o BN}} & VGG-9 & 4 & 10.00 \\
		 & & ResNet-19 & 6 / 4 / 2 & 91.98 / 90.46 / 86.87 \\
		 & \textbf{\multirow{2}{*}{IS-SNN}} & VGG-9 & 4 & \textbf{92.91} \\
		 & & ResNet-19 & 6 / 4 / 2 & \textbf{94.51 / 94.32 / 93.96} \\
		\midrule[1pt]

		\multirow{11}{*}{CIFAR-100} & Spike-Norm \cite{sengupta2019going} & VGG-16 & 2500 & 70.90 \\
		 & tdBN \cite{deng2022temporal} & ResNet-19 & 6 / 4 / 2 & 71.12 / 70.86 / 69.41 \\
		 & TET \cite{deng2022temporal} & ResNet-19 & 6 / 4 / 2 & 74.72 / 74.47 / 72.87 \\
		 & TEBN \cite{duan2022temporal} & ResNet-19 & 6 / 4 / 2 & 76.41 / 76.13 / 75.86 \\
		\cmidrule{2-5}
		 & {\multirow{2}{*}{w/o BN}} & ResNet-19 & 6 / 4 / 2 & 69.98 / 67.99 / 63.61 \\
		 & & ResNet-152 & 4 & 17.10 \\
		 & \textbf{\multirow{2}{*}{IS-SNN}} & ResNet-19 & 6 / 4 / 2 & \textbf{76.47 / 76.02 / 74.83} \\
		 & & ResNet-152 & 4 & \textbf{76.98} \\
		\midrule[1pt]

		\multirow{5}{*}{DVS-Gesture} & w/ BN \cite{fang2021deep}& 7B-Net & 16 & 97.92 \\
		 & BN-free \cite{qiao2023batch} & 3 Conv & 150 & 95.49 \\
		\cmidrule{2-5}
		 & w/o BN & 7B-Net & 16 & 77.08 \\
		 & \textbf{IS-SNN} & 7B-Net & 16 & \textbf{96.88} \\
		\midrule[1pt]

    \multirow{4}{*}{CIFAR10-DVS} & w/ BN & VGGSNN & 8 & 78.1 \\
		\cmidrule{2-5}
		 & w/o BN & VGGSNN & 8 & 10.0 \\
		 & \textbf{IS-SNN} & VGGSNN & 8 & \textbf{77.7} \\
		\midrule[1pt]

		\multirow{9.5}{*}{ImageNet} 
		 & tdBN \cite{zheng2021going} & ResNet-34 & 6 & 63.72 \\
		 & Spike-Norm \cite{sengupta2019going} & ResNet-34 & 2500 & 65.47 \\
		 & TET \cite{deng2022temporal} & ResNet-34 & 4 & 68.00 \\
		 & TEBN \cite{duan2022temporal} & ResNet-34 & 4 & 68.28 \\
		\cmidrule{2-5}
		 & w/o BN & ResNet-34 & 4 & 32.58 \\
		 & \textbf{IS-SNN} & ResNet-34 & 4 & \textbf{66.61} \\
     & \textbf{IS-SNN (E400)} & ResNet-34 & 4 & \textbf{68.05} \\
		\bottomrule[1.5pt]
	\end{tabular}
\end{table}

\begin{table}[!t]
	\caption{Performance comparison with strong data augmentation, denoted by *. With a stronger training regime, IS-SNN achieves state-of-the-art results among the compared methods, including advanced non-fusible BN variants. Its application to Spikformer further demonstrates the applicability beyond convolutional networks.}
	\label{table_com2}
  \renewcommand{\arraystretch}{0.55}
	\centering
	\begin{tabular}{p{5.5em} p{8.0em}<{\centering} p{6.0em}<{\centering} p{5.0em}<{\centering} p{11.5em}<{\centering}}
		\toprule[1.5pt]
		{Dataset} & {Model} & {Architecture} & {Timesteps} & {Accuracy (\%)} \\
		\midrule[1pt]
		\multirow{19}{*}{CIFAR-10} & TC \cite{zhou2021temporal} & VGG-16 & - & 92.68 \\
		 & RMP \cite{han2020rmp} & ResNet-20 & 2048 & 91.36 \\
		 & DSpike \cite{li2021differentiable} & ResNet-18$^*$ & 6 / 4 / 2 & 94.25 / 93.66 / 93.13 \\
		 & Spikformer \cite{zhou2022spikformer} & 4-384$^*$ & 4 & 95.19 \\
		\cmidrule{2-5}
		 & \multirow{2}{*}{TEBN \cite{duan2022temporal}} & VGG-11 & 4 & 93.96 \\
		 & & ResNet-19$^*$ & 6 / 4 / 2 & 95.60 / 95.58 / 95.45 \\
		\cmidrule{2-5}
		 & \multirow{2}{*}{TAB \cite{jiangtab}} & VGG-11 & 4 & 94.73 \\
		 & & ResNet-19$^*$ & 6 / 4 / 2 & 96.09 / 95.94 / 95.62 \\
		\cmidrule{2-5}
		 & \multirow{2}{*}{w/o BN} & VGG-11$^*$ & 4 & 10.00 \\
		 & & ResNet-19$^*$ & 6 / 4 / 2 & 92.73 / 89.94 / 85.00 \\
    \cmidrule{3-5}
		 & \textbf{\multirow{3}{*}{IS-SNN}} & VGG-11$^*$ & 4 & \textbf{95.06} \\
		 & & ResNet-19$^*$ & 6 / 4 / 2 & \textbf{96.12 / 96.02 / 95.65} \\
		 & & 4-384$^*$ & 4 & \textbf{95.43} \\
		\midrule[1pt]

		\multirow{19}{*}{CIFAR-100} & BNTT \cite{kim2021revisiting} & VGG-11 & 50 / 30 & 66.6 / 65.8 \\
		 & RMP \cite{han2020rmp} & ResNet-20 & 2048 & 67.82 \\
		 & DSpike \cite{li2021differentiable} & ResNet-18$^*$ & 6 / 4 / 2 & 74.24 / 73.35 / 71.68 \\
		 & Spikformer \cite{zhou2022spikformer} & 4-384$^*$ & 4 & 77.86 \\
		\cmidrule{2-5}
		 & \multirow{2}{*}{TEBN \cite{duan2022temporal}} & VGG-11 & 4 & 74.37 \\
		 & & ResNet-19$^*$ & 6 / 4 / 2 & 78.76 / 78.71 / 78.07 \\
		\cmidrule{2-5}
		 & \multirow{2}{*}{TAB \cite{jiangtab}} & VGG-11 & 4 & 75.89 \\
		 & & ResNet-19$^*$ & 6 / 4 / 2 & 76.82 / 76.81 / 76.31 \\
		\cmidrule{2-5}
		 & \multirow{2}{*}{w/o BN} & VGG-11$^*$ & 4 & 1.00 \\
		 & & ResNet-19$^*$ & 6 / 4 / 2 & 74.18 / 70.60 / 65.58 \\
    \cmidrule{3-5}
		 & \textbf{\multirow{3}{*}{IS-SNN}} & VGG-11$^*$ & 4 & \textbf{77.13} \\
		 & & ResNet-19$^*$ & 6 / 4 / 2 & \textbf{80.72 / 79.97 / 79.03}\\
		 & & 4-384$^*$ & 4 & \textbf{78.92} \\
		\bottomrule[1.5pt]
	\end{tabular}
\end{table}

\subsection{Experimental Setup}

Experiments were conducted on the static image classification datasets CIFAR \cite{krizhevsky2009learning} and ImageNet \cite{russakovsky2015imagenet}, as well as the neuromorphic datasets DVS-Gesture \cite{amir2017low} and CIFAR10-DVS \cite{li2017cifar10}.
All models were implemented in SpikingJelly \cite{fang2023spikingjelly} and trained with surrogate gradients, using automatic mixed precision for acceleration.
For the CIFAR datasets, models were trained for 256 epochs with an initial learning rate of 0.02 and a weight decay of 5e-4.
For ImageNet, an initial learning rate of 0.1 and a weight decay of 2e-5 were used.
Results for both 128- and 400-epoch schedules with standard augmentation techniques \cite{he2016deep} were reported to evaluate the architecture under different training budgets.
For static datasets, the standard LIF model was used.
For DVS-Gesture, the setup from 7B-Net \cite{fang2021deep} was followed.
The spike-element-wise (SEW) ResNet structure \cite{fang2021deep} was adopted for residual architectures.
Comprehensive hyperparameter configurations are provided in the Appendix.

\subsection{Performance Evaluation}

As predicted by the forward signal analysis, the IS-SNN architecture maintains stable firing rates across layers both at initialization and after training, as shown in \cref{fig:firing_rates_combined}.
This intrinsic signal stability translates into strong performance across deep architectures, as demonstrated in \cref{table_com1,table_com2}.

\Cref{table_com1} presents a systematic comparison of IS-SNN against various normalization methods.
The naive BN-free (w/o BN) models suffer severe degradation on deeper architectures: ResNet-152 on CIFAR-100 and ResNet-34 on ImageNet achieve accuracies of only 17.10\% and 32.58\%, respectively.
In contrast, IS-SNN avoids this training failure and achieves performance highly competitive with advanced, hardware-intensive dynamic BN techniques.
For ResNet-19, IS-SNN obtains substantial accuracy gains over the w/o BN baseline, with absolute improvements of 4.49\% on CIFAR-10 and 8.58\% on CIFAR-100, outperforming methods such as Spike-Norm, tdBN, and TET.
On ImageNet, the 128-epoch schedule yields 66.61\% accuracy.
Extending the training schedule to 400 epochs further improves the accuracy to 68.05\%, surpassing TET (68.00\%) and approaching the computationally expensive TEBN (68.28\%).
Considering that IS-SNN removes runtime normalization multiplications during deployment, these results indicate a favorable performance-efficiency trade-off for deep SNNs.
On the neuromorphic DVS-Gesture dataset, IS-SNN also substantially outperforms the w/o BN baseline and surpasses prior BN-free attempts.
To isolate the effect of learnable temporal dynamics, PLIF neurons were further replaced with standard LIF neurons under the same setting.
IS-SNN still achieves 95.14\%, remaining competitive with the corresponding 95.83\% w/ BN baseline.
IS-SNN was also evaluated on CIFAR10-DVS using the VGGSNN architecture, where IS-SNN achieves 77.7\%, while the standard VGGSNN without BN fails to train.

To probe the potential of IS-SNN under stronger training regimes, further experiments were conducted on the CIFAR datasets using Mixup \cite{zhang2017mixup} and Cutmix \cite{yun2019cutmix}, as detailed in \cref{table_com2}.
With ResNet-19, IS-SNN achieves state-of-the-art accuracies of 96.12\% on CIFAR-10 and 80.72\% on CIFAR-100 among the compared methods, outperforming recent advanced methods such as TAB and TEBN.
Additionally, IS-SNN can be applied to Spikformer, a Transformer-based architecture, to improve performance.
As shown in \cref{table_com2}, IS-SNN stabilizes the attention-based topology and improves performance on both CIFAR datasets, demonstrating its applicability beyond convolutional structures.

\begin{figure}[!t]
  \centering
  \begin{minipage}[]{0.49\linewidth}
    \centering
    \begin{subfigure}[b]{0.5220\linewidth}
      \includegraphics[width=\linewidth]{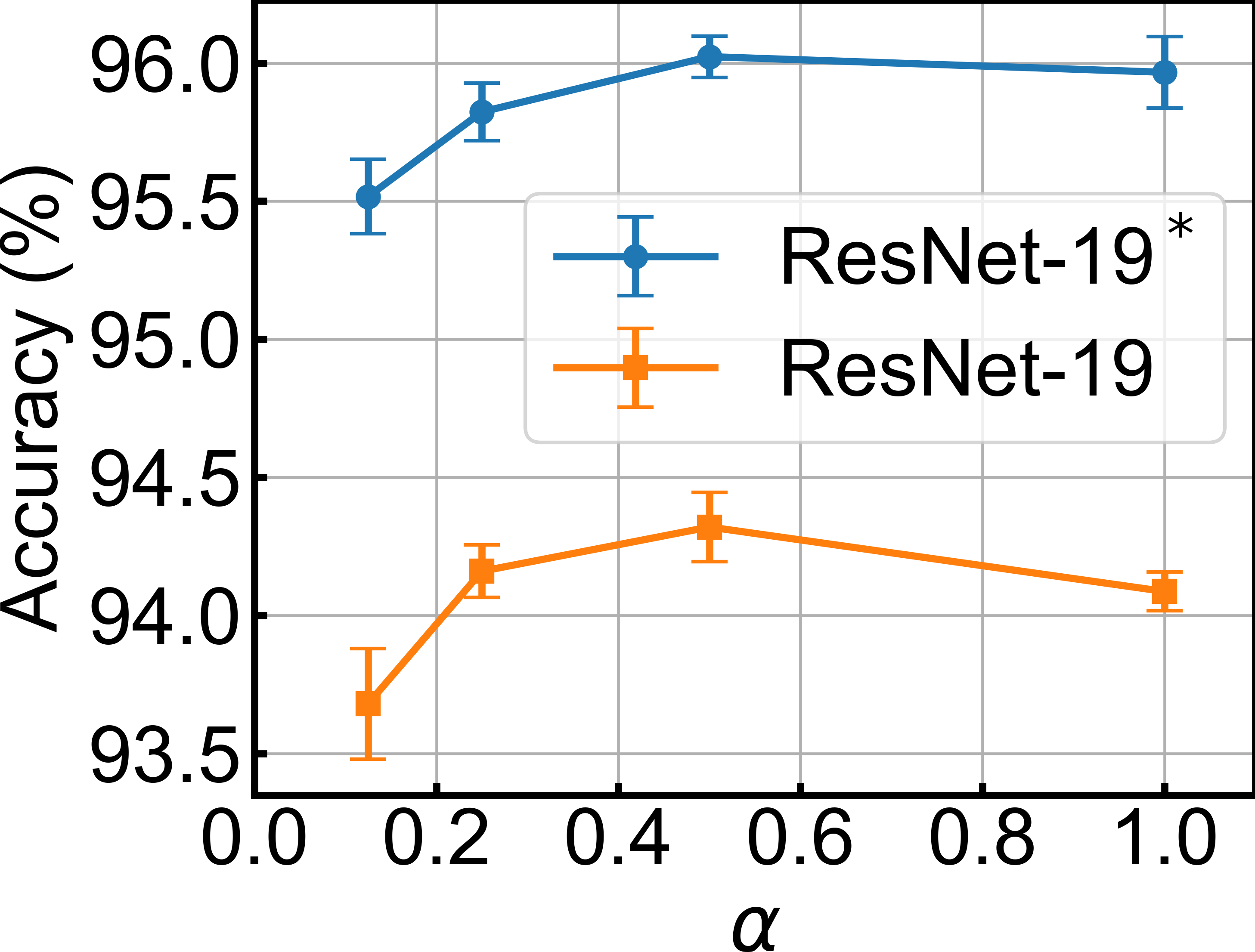}
      \caption{CIFAR-10.}
      \label{CIFAR10_a}
    \end{subfigure}
    \begin{subfigure}[b]{0.4580\linewidth}
      \includegraphics[width=\linewidth]{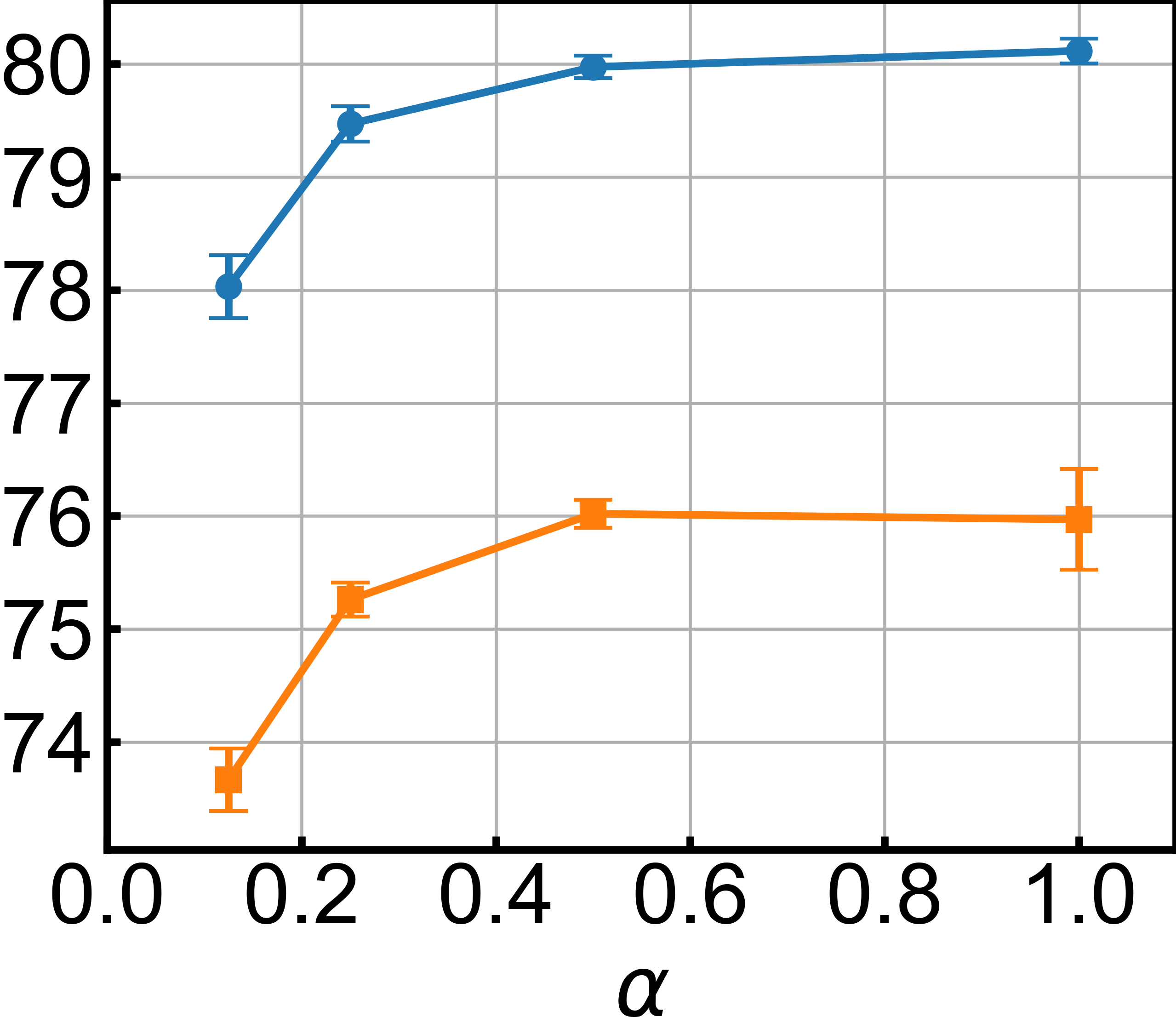}
      \caption{CIFAR-100.}
      \label{CIFAR100_a}
    \end{subfigure}
    \caption{Ablation study on the scaling factor $\alpha$. The value $\alpha=0.5$ consistently yields high accuracy with low variance across 48 total runs. The small gap between $\alpha=0.5$ and $\alpha=1.0$ indicates that the method is not highly sensitive to precise $\alpha$ tuning.}
    \label{CIFAR_a}
  \end{minipage}
  \hfill
  \begin{minipage}[]{0.49\linewidth}
    \centering
    \begin{subfigure}[b]{0.5070\linewidth}
      \includegraphics[width=\linewidth]{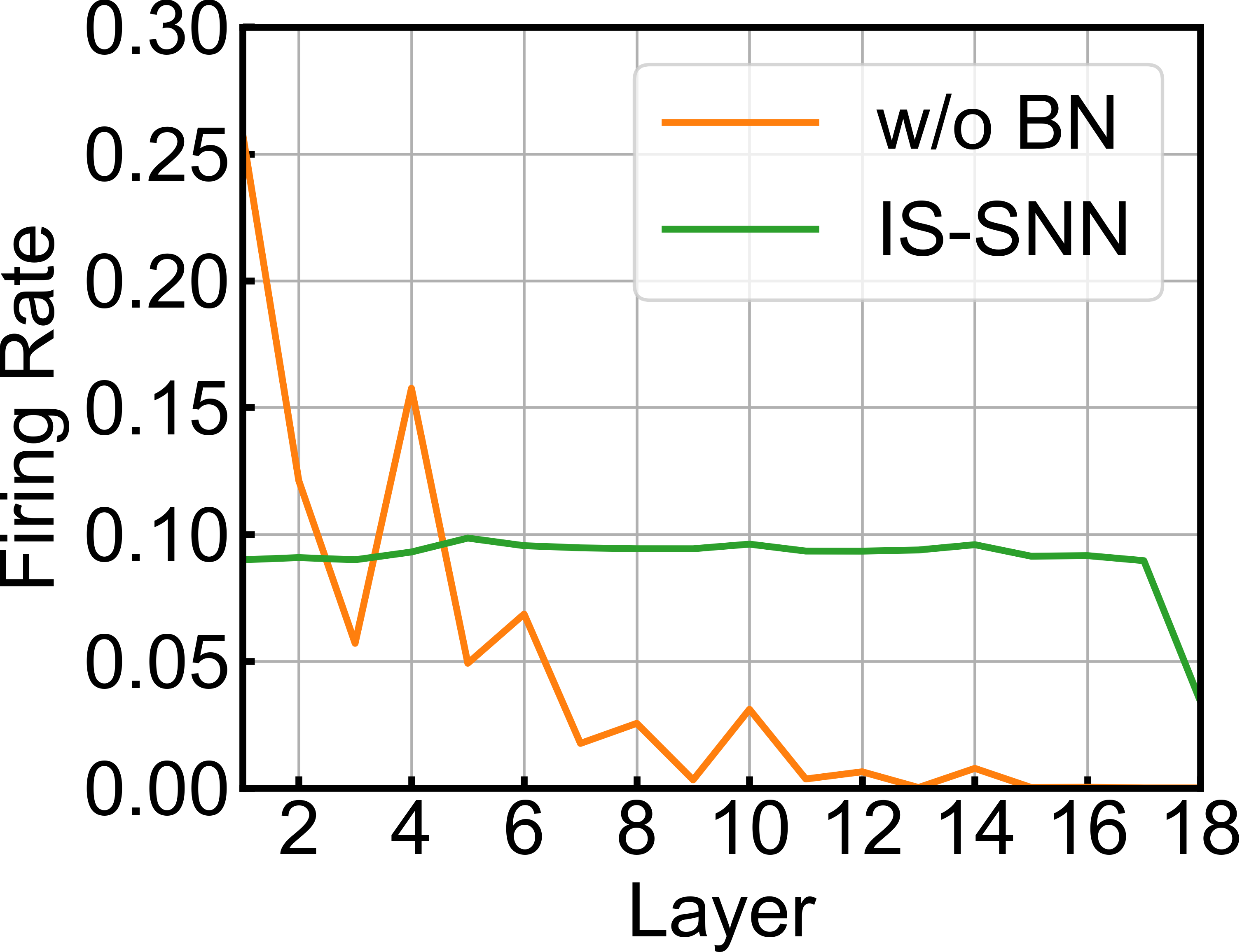}
      \caption{ResNet-19A.}
      \label{SEW19_FR_a}
    \end{subfigure}
    \begin{subfigure}[b]{0.4730\linewidth}
      \includegraphics[width=\linewidth]{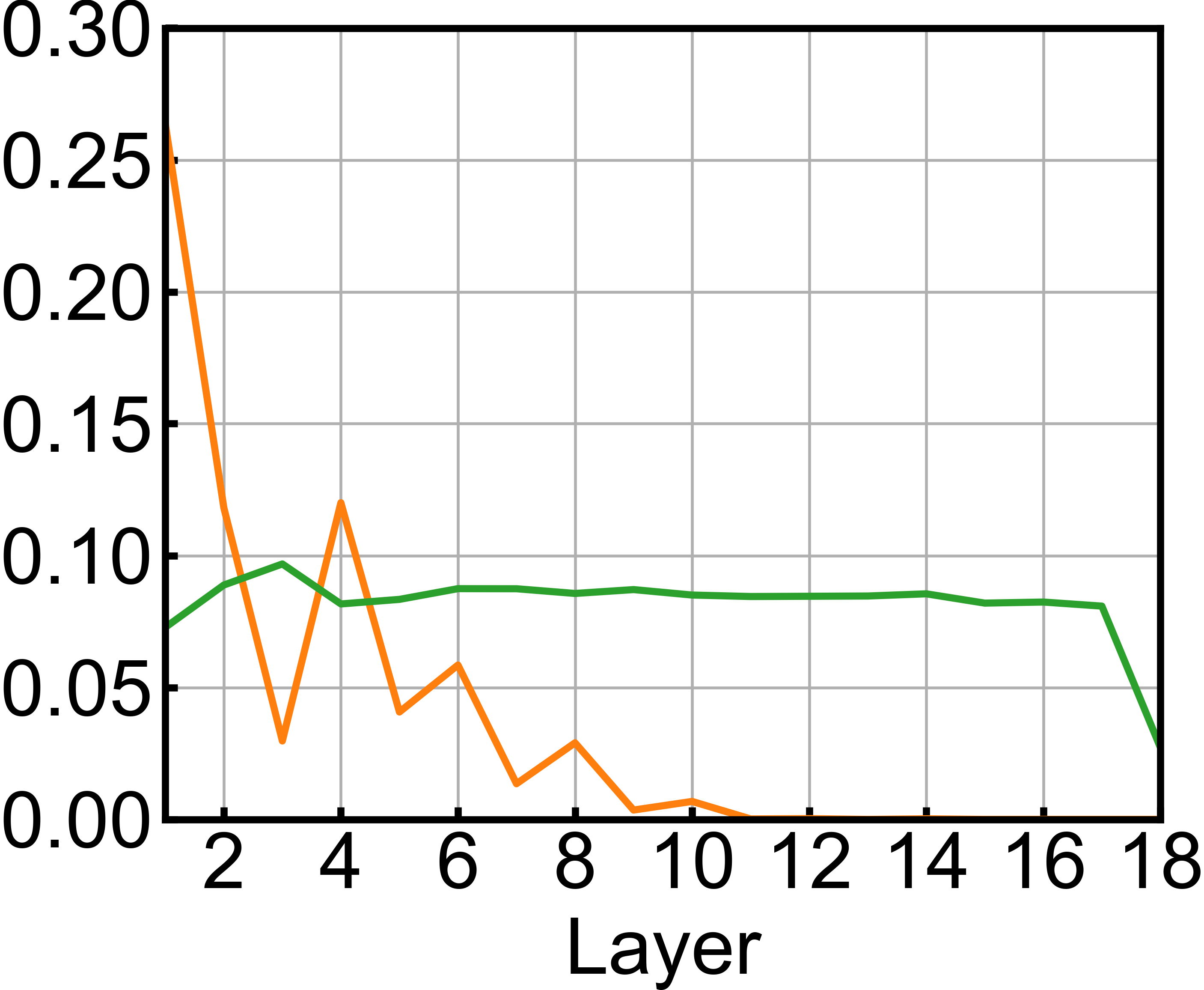}
      \caption{ResNet-19B.}
      \label{SEW19_FR_b}
    \end{subfigure}
    \caption{Initial layer-wise firing rates for two modified SEW-ResNet-19 variants on CIFAR-100. As the number of transition blocks increases, the baseline exhibits stronger rate decay, while IS-SNN remains stable across these structural variants.}
    \label{SEW19_FR}
  \end{minipage}
\end{figure}

\subsection{Ablation Study}
\label{sec:4.3}

\paragraph{Impact of the residual scaling factor $\alpha$.}
The proposed residual connection is defined as $x_{\ell+1}=\alpha\cdot$SN$(f(x_{\ell}))+x_{\ell}$, where $\alpha$ controls the variance growth rate across residual blocks.
To determine a suitable value, experiments were conducted on CIFAR datasets with $\alpha \in \{0.125, 0.25, 0.5, 1.0\}$.
These values were chosen as integer powers of 2, so the scaling operation can be implemented by an efficient bit-shift rather than standard multiplication.
To obtain statistically robust conclusions, each parameter setting was tested across three different seeds, yielding 48 full training runs in total.
As shown in \cref{CIFAR_a}, setting $\alpha=0.5$ consistently achieves strong performance with low variance.
The performance gap between $\alpha=0.5$ and $\alpha=1.0$ is small ($<0.2\%$), indicating a broad solution plateau.
This robustness allows us to adopt a unified $\alpha=0.5$ across different network backbones without exhaustive, architecture-specific hyperparameter sweeps.

\paragraph{Robustness to network structure.}
To verify whether additional transition blocks exacerbate firing-rate decay, ResNet-19 was modified to create two variants.
While all three networks share 19 layers, the standard ResNet-19 (denoted as 3C128-3C256-2C512) contains two transition blocks.
In contrast, the specially designed variants ResNet-19A (1C32-1C64-2C128-2C256-2C512) and ResNet-19B (1C32-1C64-1C96-1C128-2C256-2C512) contain four and five transition blocks, respectively.
As shown in \cref{table_as2,SEW19_FR}, networks lacking activation normalization degrade rapidly as the number of transition blocks increases, with ResNet-19B failing to converge (1.00\%).
These findings indicate that BN-free training becomes increasingly unstable as signal decay accumulates across repeated structural transitions, consistent with the behavior observed in \cref{fig:firing_rates_combined}.
In contrast, IS-SNN maintains high accuracy and stable firing rates across all topological variants, supporting its robustness to structural changes.

\begin{table}[!t]
	\caption{Accuracy comparison for the ablation study on the number of transition blocks on CIFAR-100. The w/o BN baseline degrades as the number of transition blocks increases, while IS-SNN maintains high performance across the tested variants.}
	\label{table_as2}
  \renewcommand{\arraystretch}{0.6}
	\centering
	\begin{tabular}{p{6.5em} p{6.5em}<{\centering} p{5.0em}<{\centering} p{4.5em}<{\centering} p{4.5em}<{\centering}}
		\toprule[1.5pt]
		\multirow{2.5}{*}{Architecture} & \multirow{2.5}{*}{\makecell{\# Transition \\ Blocks}} & \multirow{2.5}{*}{\# Params} & \multicolumn{2}{c}{Acc (\%)} \\
		\cmidrule{4-5}
		 & & & w/o BN & {IS-SNN} \\
		\midrule[1pt]
		ResNet-19$^*$ & 2 & 14.7 M & 70.60 & \textbf{79.97} \\
		ResNet-19A$^*$ & 4 & 13.2 M & 68.75 & \textbf{79.02} \\
		ResNet-19B$^*$ & 5 & 13.1 M & 1.00 & \textbf{78.71} \\
		\bottomrule[1.5pt]
	\end{tabular}
\end{table}

\paragraph{Naive Weight Standardization vs. IS-SNN.}
A key component of IS-SNN is the topology-aware derivation of the scaling coefficient $\gamma_\ell$.
To validate its necessity, an ablation study was conducted where $\gamma_\ell$ was fixed to $1$ for all layers.
This setting reduces the method to naive WS, which normalizes weight magnitude without explicitly addressing signal decay.
The results in \cref{table_1} show that naive WS is insufficient for deep SNNs.
It leads to severe performance degradation for ResNet and complete training collapse for VGG.
This confirms that calculating $\gamma_\ell$ based on topology-aware variance propagation is important for BN-free deep architectures.
The sensitivity of IS-SNN to possible mis-estimation of the neuron variance parameter $\sigma_g$ was further evaluated.
Using ResNet-19 on CIFAR-10/100, random perturbations were injected into $\sigma_g$ to induce deviations of up to 10\% for each layer during training.
The resulting accuracy drops were only 0.23\% and 0.31\%, respectively, indicating that IS-SNN remains robust to deviations from the estimated $\sigma_g$.

\paragraph{Synergy with Zero-Overhead Regularization.}
By eliminating dynamic BN, IS-SNN also removes the stochastic regularization effects provided by batch-wise statistics, a known trade-off for zero inference-time normalization cost.
However, because IS-SNN provides a stable baseline, it can be combined with pure training-phase stochastic regularization.
To validate this, standard Dropout was applied to ResNet-19 on CIFAR-10/100, yielding consistent accuracy gains of +0.12\% and +0.21\%, respectively.
Since Dropout is deactivated during inference, these improvements introduce no additional inference overhead.
This suggests that IS-SNN is compatible with lightweight regularization techniques, providing a way to improve accuracy without compromising inference efficiency.

\subsection{Hardware and Efficiency Benefits}

\paragraph{Training efficiency.}
Activation normalization introduces additional computational overhead during training, whether executed on server-side GPUs or through on-chip online learning \cite{xiao2022online}.
Experiments were conducted on GPUs to quantify this cost.
For ResNet-34 on ImageNet, the network with standard BN achieved a training throughput of 401 imgs/sec/GPU.
In contrast, IS-SNN achieved 461 imgs/sec/GPU, corresponding to a 15\% increase in training speed and a 17\% reduction in memory consumption.
Because advanced dynamic BN variants introduce higher overhead than standard BN, these results suggest that IS-SNN can improve training efficiency while maintaining competitive accuracy.

\paragraph{Inference efficiency.}
To evaluate inference overhead on resource-constrained hardware, deployment was simulated on an Xilinx Virtex-7 FPGA.
We compared two LIF kernels: (1) a BN-variant-enabled neuron, which requires both a multiplier and an adder for the dynamic calculations introduced by non-fusible BN variants, and (2) the proposed IS-SNN neuron.
Because IS-SNN folds the standardization operations into synaptic weights offline during compilation, its neuron kernel requires only an adder for membrane potential accumulation and a comparator for thresholding.
It therefore removes the runtime multiplication operators introduced by activation normalization.
Using standard time-division multiplexing technology \cite{pei2019towards} to compile and deploy these kernels, the IS-SNN implementation achieved a 96.4\% reduction in LUT resource consumption.
This comparison isolates the local datapath operations and does not include the additional overhead that dynamic BN variants may incur for storing and updating temporal statistics; the actual savings may therefore be greater.
Further hardware analysis and mathematical cost breakdowns are provided in the Appendix.

\begin{table}[t]
	\caption{Ablation on the scaling coefficient $\gamma_\ell$. Forcing $\gamma_\ell \equiv 1$ reduces the model to naive Weight Standardization, leading to severe degradation or complete collapse. This confirms the importance of topology-aware variance scaling for SNNs.}
	\label{table_1}
  \renewcommand{\arraystretch}{0.6}
	\centering
	\begin{tabular}{p{6.0em} p{5.8em}<{\centering} p{3.8em}<{\centering} p{3.8em}<{\centering}}
		\toprule[1.5pt]
		\multirow{2}{*}{Dataset} & \multirow{2}{*}{Structure} & \multicolumn{2}{c}{Acc (\%)} \\
		\cmidrule{3-4}
		{} & {} & {$\gamma\equiv1$} & {IS-SNN} \\
		\midrule[1pt]
		\multirow{2}{*}{CIFAR10} & {VGG-11$^*$} & 10.00 & \textbf{95.06} \\
		 & {ResNet-19} & 92.05 & \textbf{94.32} \\
		\midrule[1pt]
		\multirow{2}{*}{CIFAR100} & {VGG-11$^*$} & {1.00} & \textbf{77.13} \\
		 & {ResNet-19} & {70.69} & \textbf{76.02} \\
		\bottomrule[1.5pt]
	\end{tabular}
\end{table}
\section{Limitations and Discussion}

The IS-SNN architecture removes inference-time normalization overhead through offline reparameterization, which assumes that the weights remain fixed after training.
Thus, IS-SNN is particularly suitable for inference-focused edge deployment, whereas online learning may require additional hardware support for efficient statistics tracking.
The hardware analysis in this work focuses on the arithmetic and resource costs introduced by activation normalization, especially the runtime multiplications required by dynamic BN variants.
Although the reported FPGA results show substantial LUT savings for neuron implementations, complete end-to-end energy evaluation would also need to account for memory access, routing, and system-level scheduling.
Such full-system evaluation on dedicated neuromorphic hardware remains an important direction for future work.

Nevertheless, by removing computationally expensive normalization operations during inference, IS-SNN helps reduce the tension between high performance and hardware efficiency in deep SNNs.
This makes accurate SNN deployment on resource-constrained neuromorphic hardware more practical, especially for power-constrained edge applications where latency and energy are primary bottlenecks, such as autonomous robotics and wearable sensors.
The connection between IS-SNN and homeostatic regulation also suggests that neuroscience-inspired stability principles can provide useful guidance for efficient systems.

This work opens several promising avenues for future research.
First, the stable signal-propagation behavior provided by IS-SNN can serve as a baseline for investigating trade-offs among firing rate, spike sparsity, and task performance in SNNs.
Second, future research could adapt these intrinsic stabilization mechanisms to hardware-friendly online learning paradigms.
Third, extending IS-SNN beyond vision benchmarks to sequence-centric tasks and to unsupervised or self-supervised learning may further test the generality of BN-free SNNs.

\section{Conclusion}
\label{sec:Conclusion}

This work addresses the tension between performance and hardware efficiency in deep spiking neural networks, which arises from the reliance on computationally expensive, non-fusible BN variants.
Building on the observation that catastrophic firing-rate decay is a primary cause of signal collapse in networks lacking normalization, the proposed IS-SNN architecture removes activation-normalization layers by enforcing intrinsic signal stability through topology-aware weight standardization and offline reparameterization.
Comprehensive experiments show that IS-SNN achieves performance competitive with or superior to advanced dynamic BN techniques across diverse network topologies, including VGG, ResNet, and Transformers, on benchmarks ranging from CIFAR to ImageNet.
Its effectiveness in deep architectures is further supported by a 35.47\% accuracy improvement on ImageNet over a naive BN-free baseline.
By delivering highly competitive accuracy together with no inference-time normalization overhead and a 96.4\% reduction in FPGA LUT resource consumption for neuron implementations, this work provides a practical framework for designing SNNs that are accurate, hardware-friendly, and biologically motivated.

\section*{Acknowledgements}
This work was supported in part by the Brain Science and Brain-like Intelligence Technology-National Science and Technology Major Project 2022ZD0209700 and in part by Sichuan Provincial Industrial Development Fund under Grant 2025JB0\allowbreak4 and in part by Shenzhen Natural Science Foundation under Grant JCYJ202506\allowbreak04180428038.

%
%
{
\bibliographystyle{splncs04}
\bibliography{main}
}

\clearpage
\appendix

\title{Supplementary Material for \\ Intrinsically Stable Spiking Neural Networks: Overcoming the Performance Barrier in the Absence of Batch Normalization}

\titlerunning{IS-SNN}

\author{Ruichen Ma\inst{1}\orcid{0009-0004-6210-8022} \and
Xiaoyang Zhang\inst{2}\orcid{0009-0005-8084-2906} \and
Jian Bai\inst{2}\orcid{0000-0002-9845-6991} \and
Guanchao Qiao\inst{1}\orcid{0000-0003-4982-5938} \and
Liwei Meng\inst{1}\orcid{0009-0002-0842-1224} \and
Ning Ning\inst{1} \and
Yang Liu\inst{1} \and
Shaogang Hu\inst{1,3}$^*$\orcid{0000-0002-8653-2491}}

\authorrunning{Ma et al.}

\institute{University of Electronic Science and Technology of China (UESTC) \and
Beijing Institute of Remote-Sensing Equipment \and
Shenzhen Institute for Advanced Study, UESTC \\
\url{https://github.com/Ruichen0424/IS-SNN}
}

\maketitle

\section{Spiking Neural Networks}
\begin{wrapfigure}{r}{0.50\textwidth}
  \centering
  \includegraphics[width=\linewidth]{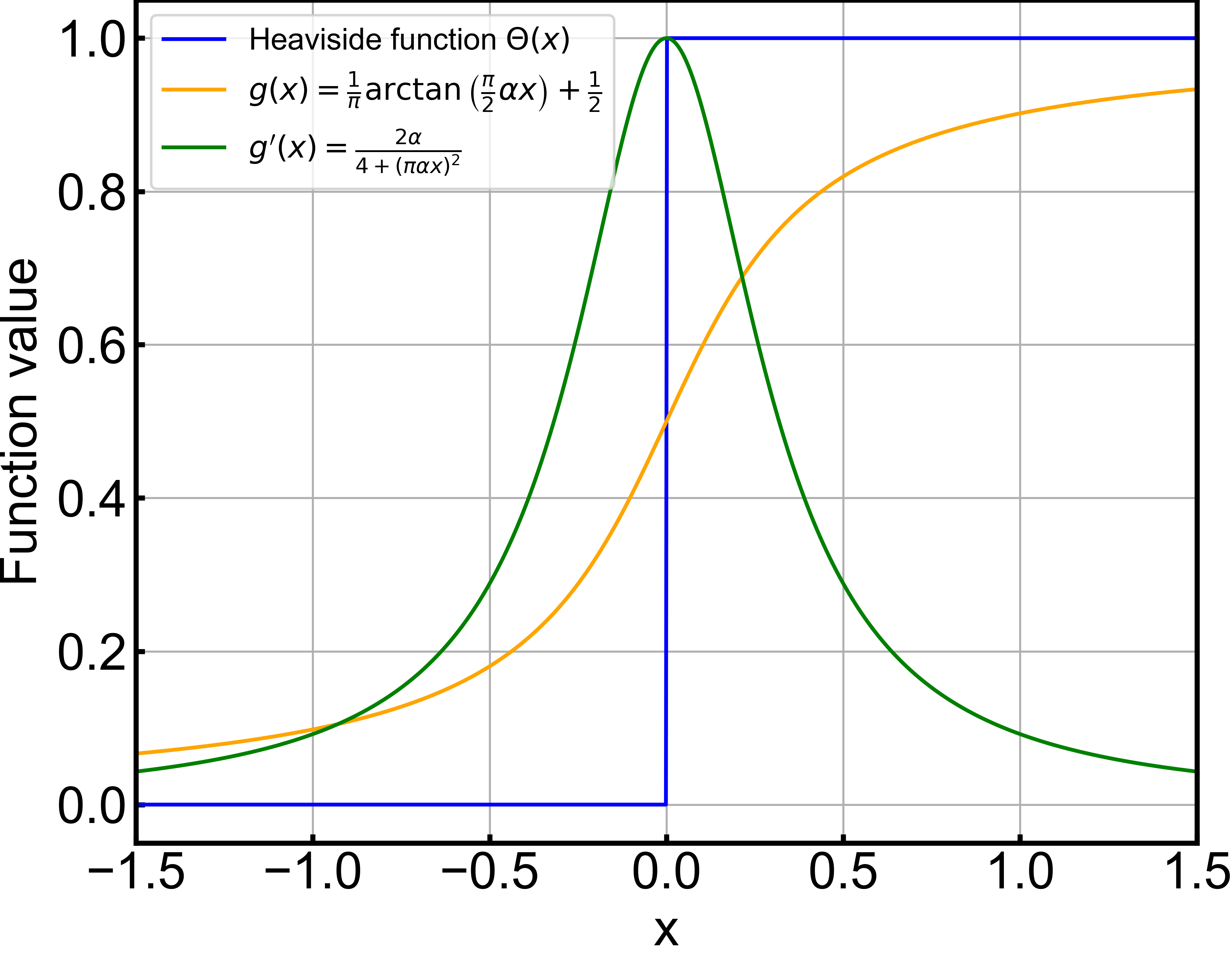}
  \caption{The surrogate gradient method in SNN training.
  The non-differentiable Heaviside step function $\Theta(x)$ is used for spike generation in the forward pass.
  It is approximated by a smooth surrogate function $g(x)$, whose well-defined derivative $g'(x)$ is used in the backward pass.}
  \label{figs1}
\end{wrapfigure}

A fundamental challenge in training SNNs is the non-differentiable nature of the spike activation function.
The neuronal firing process is typically modeled by the Heaviside step function $\Theta(\cdot)$.
Because the derivative of $\Theta(\cdot)$ is zero almost everywhere and mathematically undefined at the firing threshold, standard gradient-based optimization cannot be directly applied.
To address this issue, we employ the surrogate gradient method.
This technique replaces the true, ill-defined gradient of the Heaviside function with the gradient of a smooth, differentiable surrogate during the backward pass.
In this work, we use an arctan-based surrogate function, $g(x)$, which provides a continuous approximation of the step function:
\begin{equation}
g(x) = \frac{1}{\pi} \arctan\left(\frac{\pi}{2}\alpha x\right) + \frac{1}{2}
\end{equation}
The corresponding gradient, which is substituted for the true gradient, is:
\begin{equation}
g'(x) = \frac{2\alpha}{4+(\pi\alpha x)^2}
\end{equation}
The hyperparameter $\alpha$ controls the steepness of the surrogate function and modulates the sharpness of the approximated gradient.
A larger $\alpha$ gives a closer approximation to the Heaviside function but creates a more localized and narrower gradient landscape.
We set $\alpha=2$ in this work.
\Cref{figs1} illustrates the relationship between the Heaviside function, the chosen surrogate, and its gradient.

\section{Empirical Estimation of Output Variance $\sigma_g^2$}

\begin{figure}[!t]
  \centering
  \begin{subfigure}{0.342\linewidth}
    \centering
    \includegraphics[width=\linewidth]{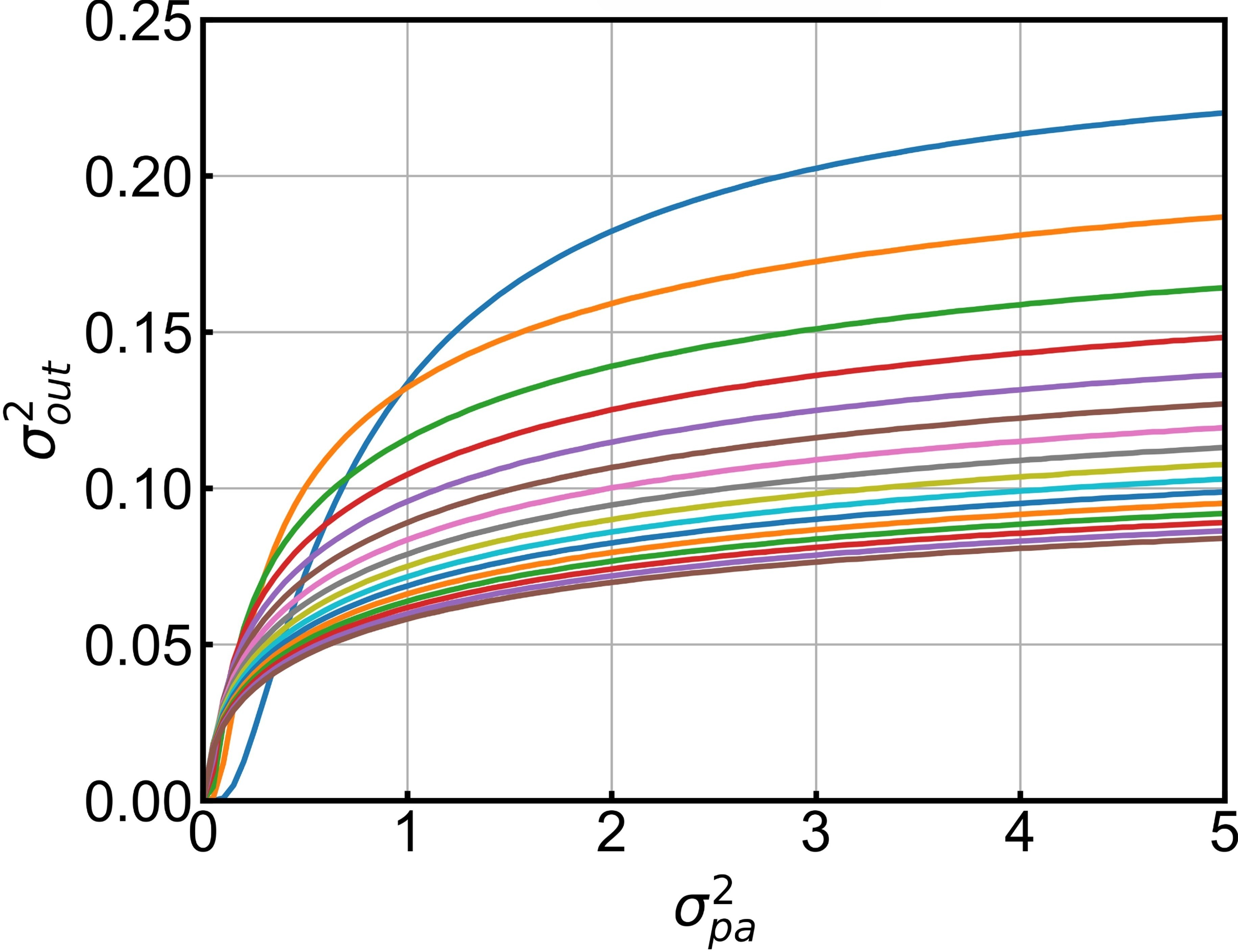}
    \caption{IF model.}
    \label{figs2a}
  \end{subfigure}
  \begin{subfigure}{0.318\linewidth}
    \centering
    \includegraphics[width=\linewidth]{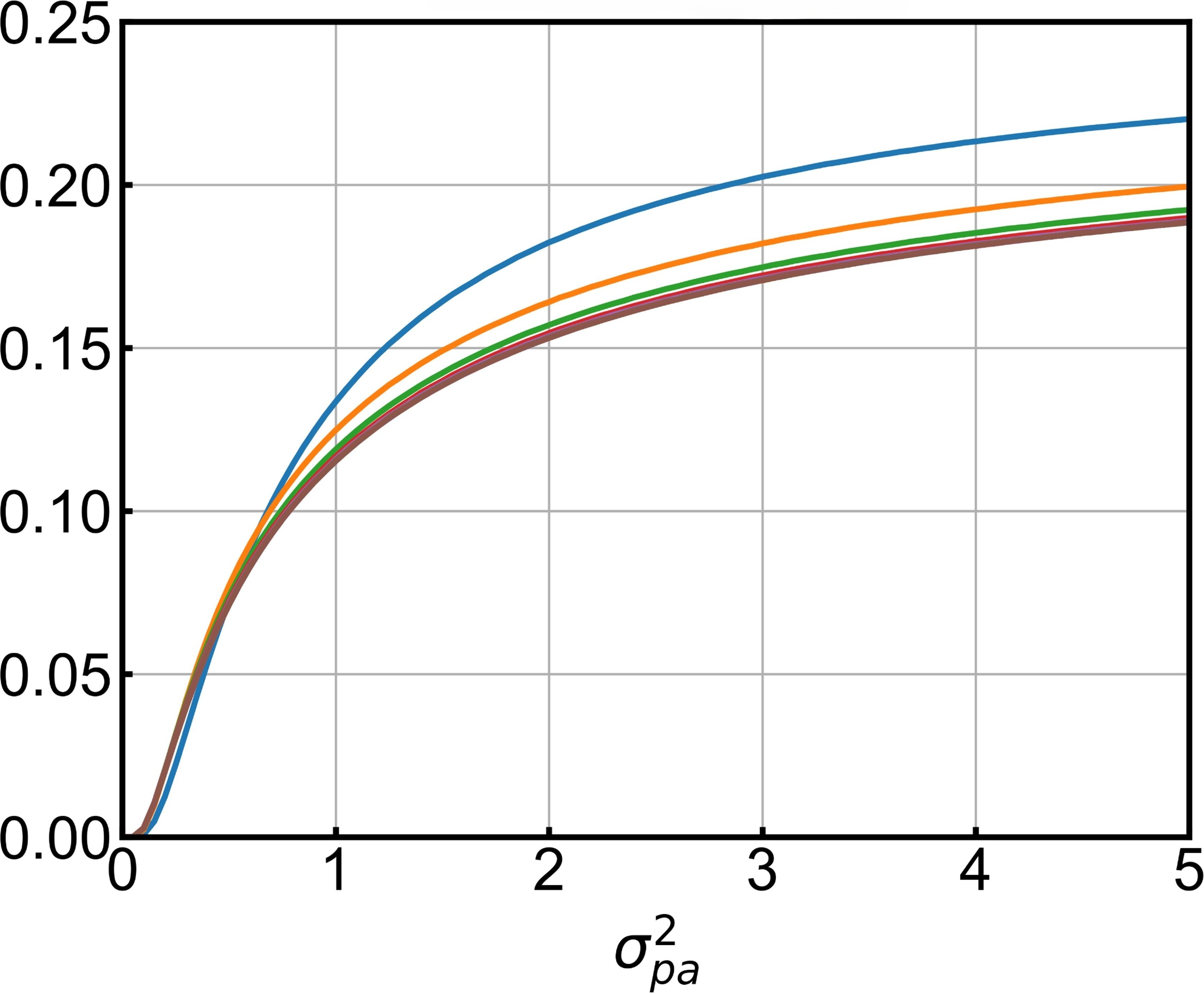}
    \caption{LIF model ($\tau=2$).}
    \label{figs2b}
  \end{subfigure}
  \begin{subfigure}{0.318\linewidth}
    \centering
    \includegraphics[width=\linewidth]{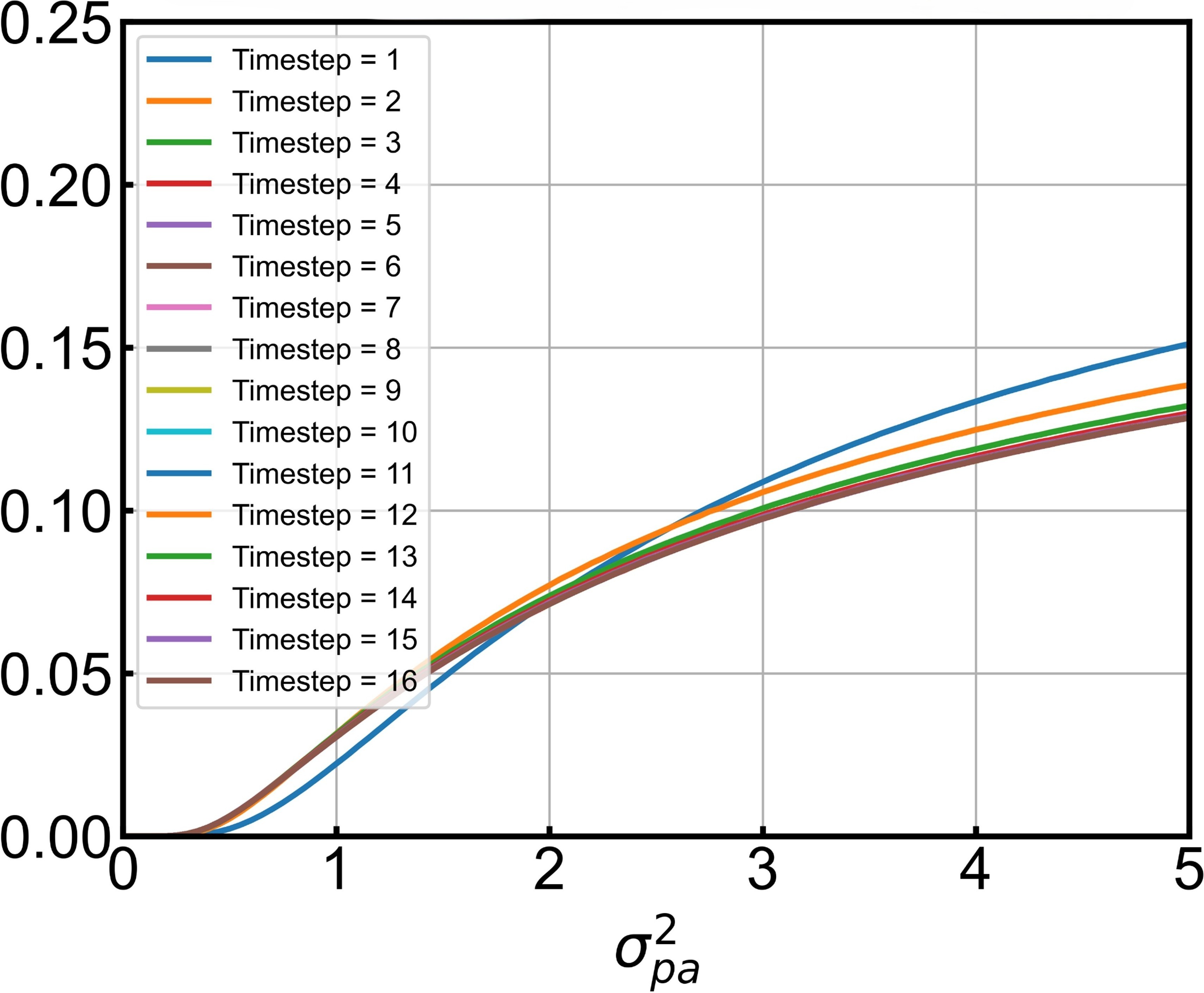}
    \caption{LIF model with decaying.}
    \label{figs2c}
  \end{subfigure}
  \caption{Empirical output spike variance $\sigma_g^2$ as a function of the pre-activation variance $\sigma_{pa}^2$ for different neuron models.
  The pre-activations are drawn from a zero-mean Gaussian distribution.
  These plots show how different neuronal dynamics, including non-leaky and leaky integration, affect the statistics of the output spike train.}
  \label{figs2}
\end{figure}

As established in the main text, the construction of IS-SNN depends on characterizing the base output spike variance $\sigma_g^2$ for specific neuron models.
To this end, this value is empirically estimated through simulations where the pre-activation $x_{pa}$ is modeled as a random variable drawn from a Gaussian distribution, $x_{pa} \sim \mathcal{N}(\mu_{pa}, \sigma_{pa}^2)$.
The analysis focuses on the zero-mean case ($\mu_{pa}=0$), where the input variance $\sigma_{pa}^2$ is systematically varied, and the resulting output variance is measured.

\Cref{figs2} presents the simulation results.
For the standard IF neuron (\cref{figs2a}), the absence of a membrane potential leak ($\tau \to \infty$) means that the potential integrates incoming inputs without decay.
Consequently, after receiving a random continuous sequence of inputs, the average membrane potential tends to decrease below the firing threshold.
This occurs because positive potential accumulations are capped and reset upon firing, whereas negative potential accumulations have no such lower bound and can continue to decrease.
This asymmetric accumulation leads to a progressively lower firing rate over time.
In contrast, the LIF neuron (\cref{figs2b}), configured with a membrane time constant $\tau=2$, includes a continuous decay mechanism that pulls the membrane potential back toward its resting state, thereby promoting more regular and sustained firing patterns.
Furthermore, the LIF model with decaying input currents (\cref{figs2c}) exhibits a stronger reduction in overall firing activity.
The attenuation of the input current means that less charge is integrated per timestep, making it harder for the neuron to reach its firing threshold and resulting in a lower output variance.

The estimation is anchored at a pre-activation variance of $\sigma_{pa}^2=1$.
This choice corresponds to the reference state induced by BN.
Because BN normalizes pre-activations toward a standard normal distribution $\mathcal{N}(0, 1)$, measuring the neuron output variance under this condition allows IS-SNN to approximate the signal scale of a BN-equipped network.

Finally, the scaling parameters used by IS-SNN are static for a given architecture.
The layer-wise expected input variance $\sigma_\ell^2$ is derived from network topology, rather than being tracked at runtime.
Consequently, the base variance $\sigma_g^2$, the expected input variance $\sigma_\ell^2$, and the resulting scaling factor $\gamma_\ell$ are pre-calculated constants for each architecture.
They remain fixed during both training and inference.
The only quantities that fluctuate during training are the batch-wise mean and variance of the learnable weights ($\mu_i, \sigma_i^2$).
Once training is complete and the model is deployed for inference, these weight statistics are fixed, enabling the standardization operations to be folded into weights offline.

\section{Experimental Details}

\subsection{Experimental Setup}
All experiments were conducted on a local Ubuntu 22.04 server equipped with two NVIDIA RTX 5090 GPUs.
We used the SpikingJelly framework for all model implementations and training pipelines.
To ensure reproducibility, the random seed for all stochastic processes was fixed to $2024$.
The only exception is the robustness ablation study detailed in Sec. 4.3 of the main text, where three distinct seeds, $2024$, $2025$, and $2026$, were used to calculate the statistical variance of model accuracy.

\begin{table}[t]
    \caption{Hyperparameters for training IS-SNN models from scratch.}
    \label{table_hyper1}
    \centering
    \begin{tabular}{p{6.5em} p{6.7em}<{\centering} p{6.9em}<{\centering} p{5.4em}<{\centering} p{6.5em}<{\centering}}
        \toprule[1.5pt]
        {Dataset} & {Learning Rate} & \makecell{Weight Decay} & {Batch Size} & {Epochs}\\
        \midrule[1pt]
        \multirow{1}{*}{CIFAR} & 0.02 & 5e-4 & 128 & 256\\
        \multirow{1}{*}{ImageNet} & 0.1 & 2e-5 & 256 & 128 / 400\\
        \multirow{1}{*}{DVS-Gesture} & 0.01 & 5e-4 & 16 & 196\\
        \bottomrule[1.5pt]
    \end{tabular}
\end{table}

\subsection{Training Configuration and Neuron Parameters}
\label{sec:training_config}
For all convolutional architectures (VGG and ResNet), we employed the SGD optimizer with a momentum of $0.9$.
A cosine annealing schedule was applied to decay the learning rate.
For the Spikformer architecture, we adopted the AdamW optimizer, following standard practice for Transformer-based models.

Across all experiments and variance estimations, the firing threshold was set to $V_{th} = 1.0$, and the resting membrane potential was initialized and reset to $V_{rest} = 0.0$.
Following established practices in SNN training, the gradient flow through the neuron's hard reset mechanism was detached during the backward pass to improve optimization stability.
The hyperparameters used for each dataset are detailed in \cref{table_hyper1}.

\begin{figure}[!t]
  \centering
  \begin{subfigure}{0.2572\linewidth}
    \centering
    \includegraphics[width=\linewidth]{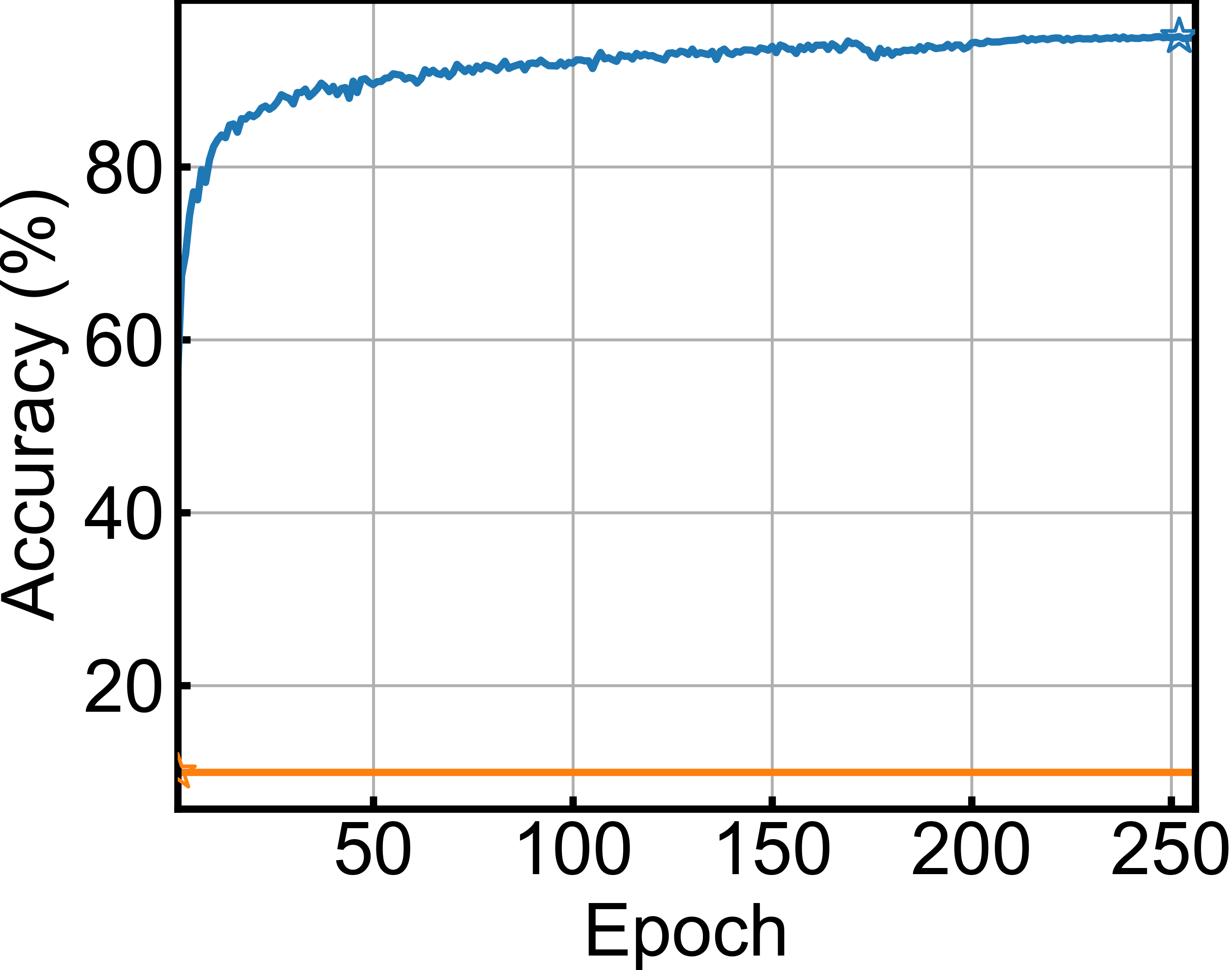}
    \caption{VGG-11$^*$ on CIFAR-10.}
    \label{figs3a}
  \end{subfigure}
  \begin{subfigure}{0.2395\linewidth}
    \centering
    \includegraphics[width=\linewidth]{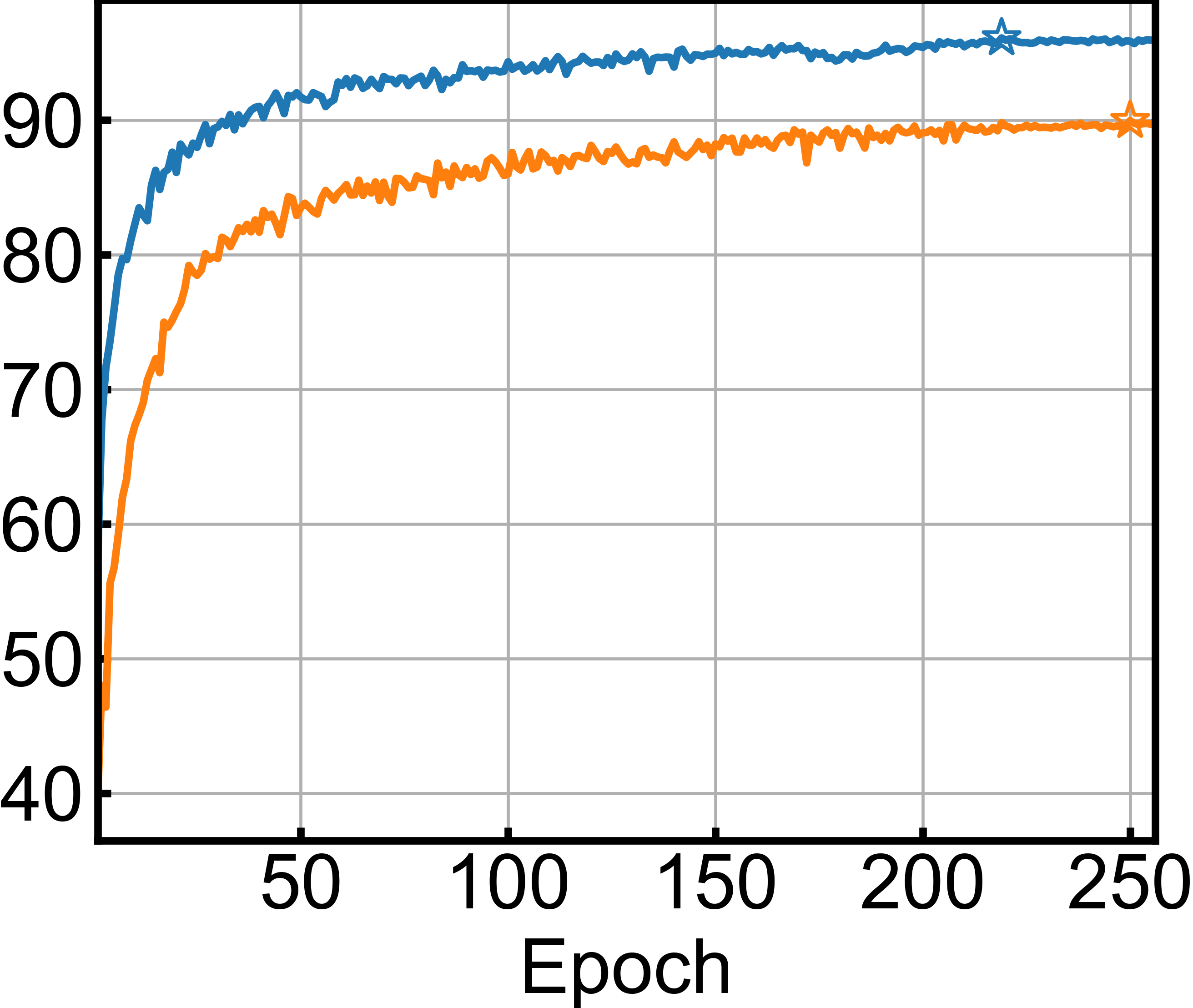}
    \caption{ResNet-19$^*$ on CIFAR-10.}
    \label{figs3b}
  \end{subfigure}
  \begin{subfigure}{0.2395\linewidth}
    \centering
    \includegraphics[width=\linewidth]{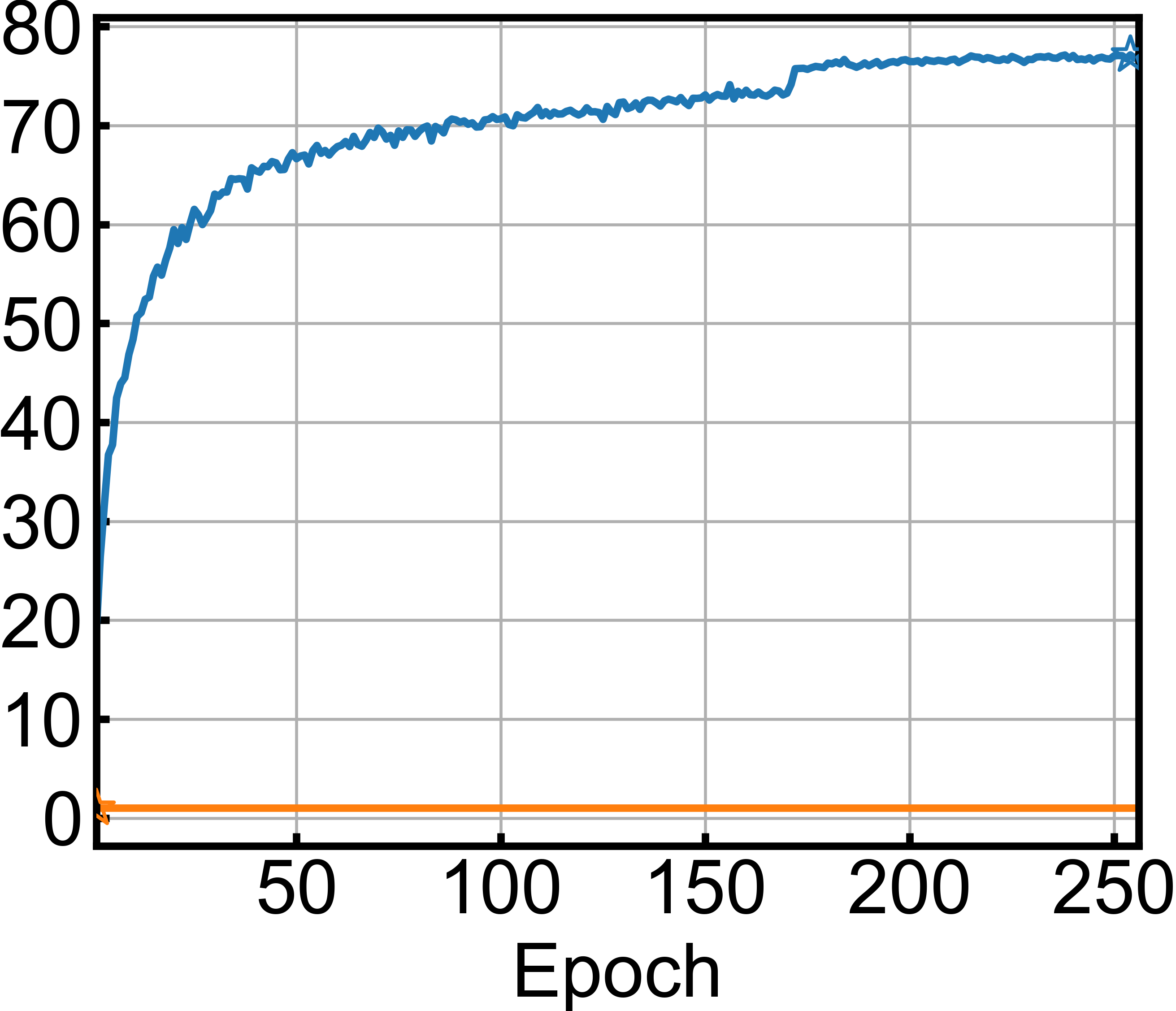}
    \caption{VGG-11$^*$ on CIFAR-100.}
    \label{figs3c}
  \end{subfigure}
  \begin{subfigure}{0.2397\linewidth}
    \centering
    \includegraphics[width=\linewidth]{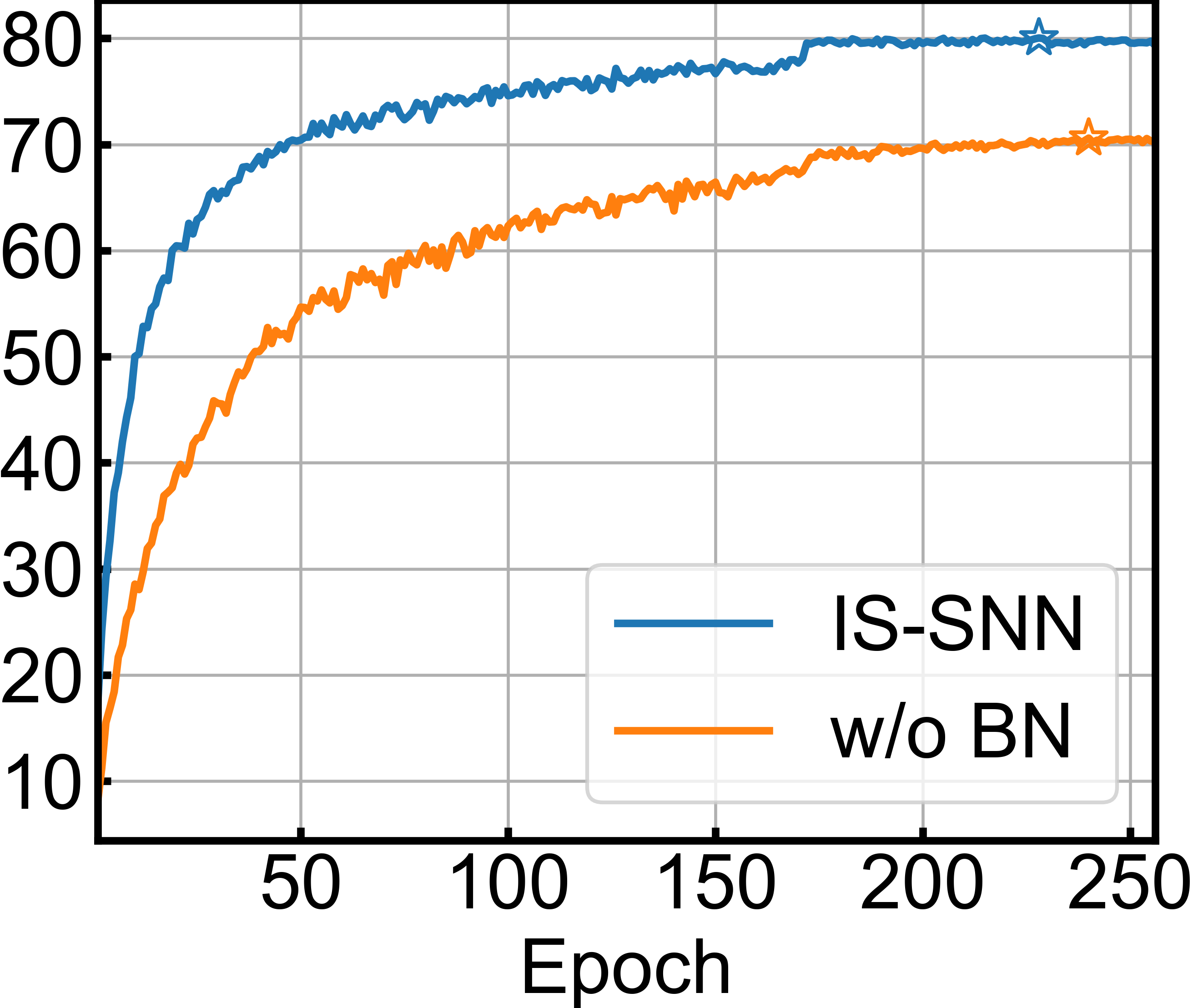}
    \caption{ResNet-19$^*$ on CIFAR-100.}
    \label{figs3d}
  \end{subfigure}
  \caption{Validation accuracy on CIFAR-10 and CIFAR-100. The naive w/o BN baseline fails to converge on VGG, while IS-SNN maintains stable training.}
  \label{figs3}
\end{figure}

\begin{wrapfigure}{r}{0.50\textwidth}
  \centering
  \begin{subfigure}{0.4991\linewidth}
    \centering
    \includegraphics[width=\linewidth]{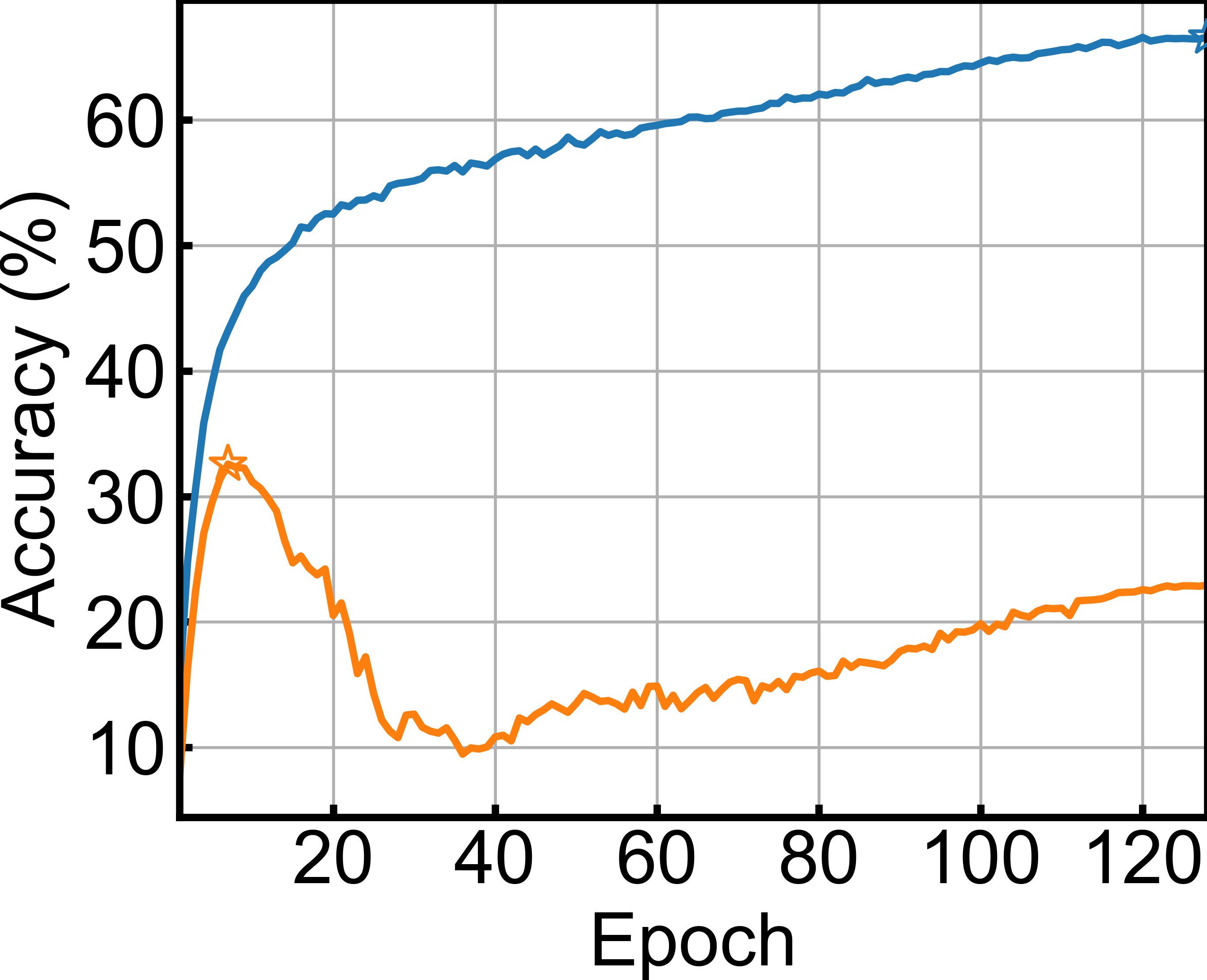}
    \caption{ImageNet.}
    \label{figs4a}
  \end{subfigure}
  \begin{subfigure}{0.4810\linewidth}
    \centering
    \includegraphics[width=\linewidth]{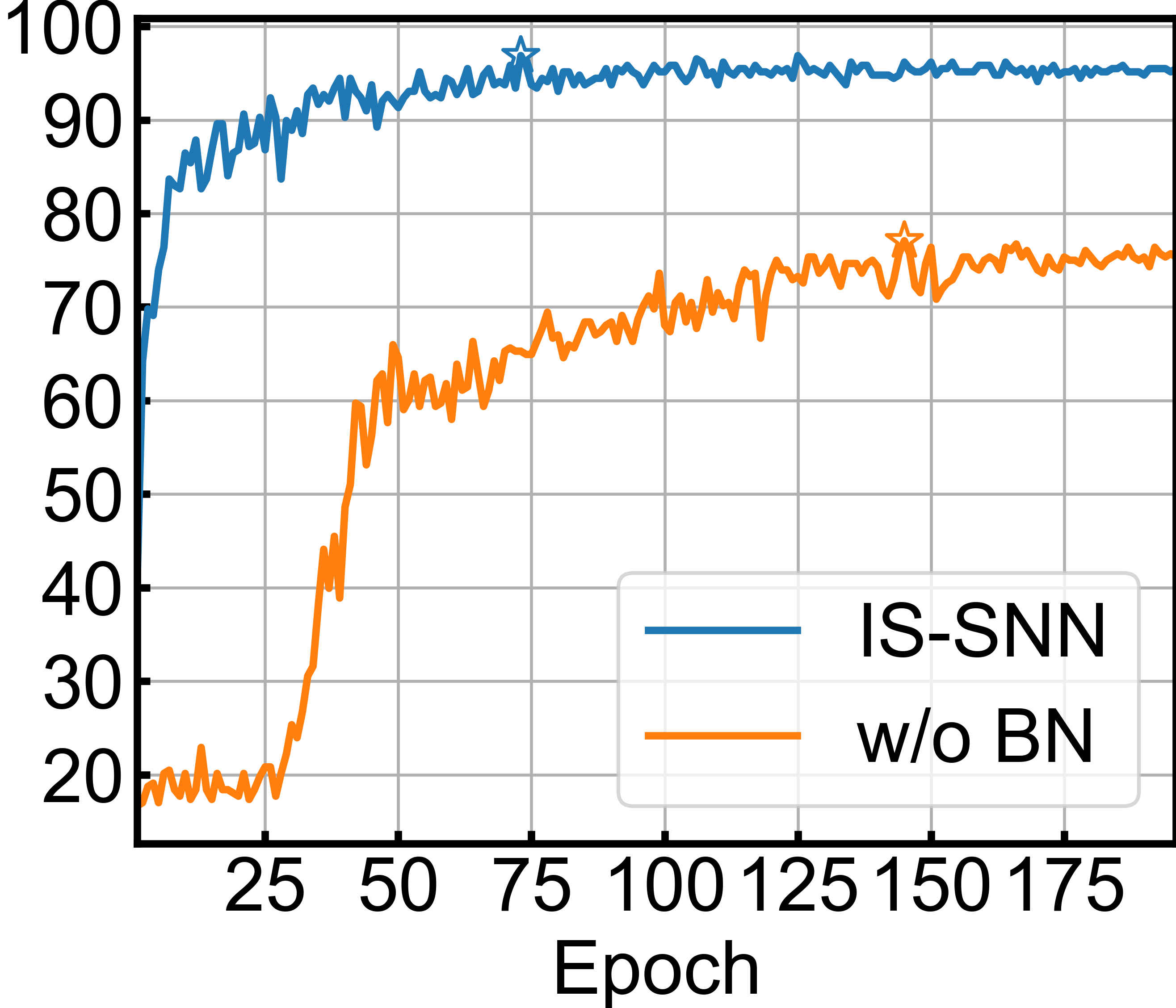}
    \caption{DVS-Gesture.}
    \label{figs4b}
  \end{subfigure}
  \caption{Validation accuracy curves on ImageNet and DVS-Gesture. IS-SNN prevents the training failures observed in the w/o BN baselines.}
  \label{figs4}
\end{wrapfigure}

\subsection{Data Augmentation}
Different data augmentation strategies were applied according to the nature of each dataset.

\textbf{CIFAR-10/100:}
Standard random horizontal flips and random crops were used as the baseline augmentation.
For experiments denoted with strong augmentation (*), either Mixup or Cutmix ($\alpha=1.0$) was randomly applied to each mini-batch during the first two-thirds of the total training epochs.

\textbf{ImageNet:}
Standard augmentations were employed, primarily consisting of random resized crops and horizontal flips.

\textbf{DVS-Gesture:}
The random temporal deletion strategy was used.
Specifically, during training, $12$ out of the $16$ available timesteps were randomly selected for each input sample to prevent temporal overfitting.
During inference, the full, un-augmented $16$ timesteps were used to evaluate the final performance.

For experiments involving the Spikformer architecture, the same data augmentation scheme as reported in the original Spikformer work was adopted for fair comparison.

\subsection{Analysis of Training Dynamics}
\Cref{figs3,figs4} illustrate the training dynamics of IS-SNN compared with baseline models without batch normalization.

As shown in \cref{figs3}, removing BN from the VGG-11 architecture on the CIFAR datasets leads to failure to converge, with accuracy remaining near chance level.
Although the ResNet-19 architecture exhibits greater structural robustness, its performance remains substantially degraded in the absence of BN.
In contrast, IS-SNN maintains stable training across the tested architectures and datasets, achieving strong validation accuracy.

This instability of BN-free networks becomes more pronounced on complex datasets.
On ImageNet (\cref{figs4a}), the absence of BN's adaptive normalization leads to numerical issues early in training, such as exploding or vanishing gradients, preventing the model from learning effective representations.
Similarly, on DVS-Gesture (\cref{figs4b}), the naive BN-free baseline fails to converge for many early epochs and suffers a substantial performance drop.
By contrast, IS-SNN maintains stable training dynamics and achieves final validation results competitive with BN-equipped counterparts, supporting its effectiveness in regulating signal propagation in deep SNNs.

\section{Detailed Analysis of Hardware Efficiency}
\label{sec:hardware_analysis}

This section provides a theoretical breakdown of the hardware efficiency gains reported in the main text.
We analyze the non-fusibility of dynamic BN variants used in state-of-the-art SNNs and provide a resource estimation based on FPGA logic synthesis principles.

\subsection{Non-Fusibility of Dynamic BN Variants}
\label{subsec:non_fusibility}

In conventional ANNs, BN is typically applied after a convolutional layer and before the activation function.
Let $x_{in}$ be the input to a convolutional layer with weights $W$ and bias $b_{conv}$.
The output of this layer is $x = W * x_{in} + b_{conv}$.
The standard static BN operation during inference is defined as:
\begin{equation}
    \text{BN}(x) = \gamma \cdot \frac{x - \mu_{\text{run}}}{\sqrt{\sigma_{\text{run}}^2 + \epsilon}} + \beta
\end{equation}
where $\mu_{\text{run}}$ and $\sigma_{\text{run}}^2$ denote the running mean and variance accumulated during training, and $\gamma, \beta$ are the learnable affine parameters.
Because $\mu_{\text{run}}$, $\sigma_{\text{run}}$, $\gamma$, and $\beta$ are fixed during inference, the BN layer can be mathematically folded into the preceding convolutional layer.
The combined sequential operation transforms from:
\begin{equation}
    y = \text{BN}(W * x_{in} + b_{conv})
\end{equation}
to a fused convolution:
\begin{equation}
    y = W_{\text{fused}} * x_{in} + b_{\text{fused}}
\end{equation}
where the fused parameters are pre-computed offline as:
\begin{equation}
    W_{\text{fused}} = \frac{W \cdot \gamma}{\sqrt{\sigma_{\text{run}}^2 + \epsilon}}, \quad b_{\text{fused}} = \frac{\gamma \cdot (b_{conv} - \mu_{\text{run}})}{\sqrt{\sigma_{\text{run}}^2 + \epsilon}} + \beta
\end{equation}
This reparameterization removes the activation-normalization operation during inference.

However, to mitigate firing-rate decay in deep SNNs, recent methods (\eg, TEBN and tdBN) introduce time-dependent or batch-dependent dynamic normalization.
In these variants, the statistics depend on the runtime timestep $t$:
\begin{equation}
    y(t) = \gamma(t) \cdot \frac{x(t) - \mu(t)}{\sqrt{\sigma(t)^2 + \epsilon}} + \beta(t)
\end{equation}
Because the scaling factor $\frac{\gamma(t)}{\sigma(t)}$ and the shift term vary across timesteps or batches, they cannot be absorbed into fixed static weights $W$.
Consequently, the hardware must execute floating-point or fixed-point multiplications and additions for normalization during inference.
Furthermore, even if a standard static BN layer is fused, the resulting fused bias term, $b_{\text{fused}}$, can be undesirable for neuromorphic implementations.
Neuromorphic hardware relies on event-driven sparsity, where compute operations are triggered by incoming spikes.
A non-zero bias term injects a constant background current into the membrane potential at every timestep, independent of input spike activity.
This can force membrane updates and reduce the effective sparsity of the network.
By eliminating activation normalization, IS-SNN avoids both the non-fusible normalization multiplications and the constant bias injection introduced by fused BN.

\subsection{Computational Complexity Breakdown}

The computational overhead introduced by non-fusible dynamic BN variants can be categorized into two components:
\begin{itemize}
    \item \textbf{Component A: Dynamic Statistics Aggregation.} Computing the dynamic mean $\mu(t)$ and variance $\sigma^2(t)$ requires global summation or sliding average operations across the feature map. This involves memory access bandwidth and aggregation routing logic.
    \item \textbf{Component B: Affine Transformation.} Applying normalization dynamically requires element-wise multiply-accumulate (MAC) operations ($x \cdot A(t) + B(t)$) for every pre-activation vector.
\end{itemize}
In the FPGA hardware analysis presented in Sec. 4.4, only the resource savings derived from eliminating Component B were quantified.
This assumes a simplified scenario where the dynamic statistics are already available, and the comparison focuses on the arithmetic logic required for the local neuron datapath.
Because IS-SNN removes both dynamic statistics aggregation and dynamic affine normalization during inference, full-system savings may be larger than the local datapath savings reported here.

\subsection{FPGA Resource Estimation Analysis}
\label{subsec:fpga_estimation}

We further clarify the source of the 96.4\% reduction in Look-Up Table (LUT) resources reported in the hardware experiments.
The analytical comparison is based on a standard 32-bit precision digital implementation deployed on an Xilinx Virtex-7 FPGA.

\textbf{Resource Modeling:}
In digital FPGA architectures, the base resource cost of fundamental arithmetic units differs substantially:
\begin{itemize}
    \item \textbf{Adders:} An $N$-bit adder is area-efficient, typically consuming $\mathcal{O}(N)$ LUTs (approximately $N$ LUTs) by using dedicated fast carry-chain logic inside DSP slices or logic cells.
    \item \textbf{Multipliers:} A combinational $N$-bit $\times$ $N$-bit multiplier is more resource-intensive. Its logic complexity scales quadratically, consuming $\mathcal{O}(N^2)$ LUTs (empirically about $0.5 N^2$ LUTs for standard parallel array synthesis).
\end{itemize}

\textbf{Theoretical Reduction:}
The datapath of a dynamic BN-equipped neuron, which requires runtime multiplication for scaling, is compared with the proposed IS-SNN neuron, which requires only accumulation after offline folding.
Assuming a standard datapath width of $N=32$:
\begin{itemize}
    \item \textbf{BN-equipped Neuron:} At least one multiplier is required to apply the dynamic scaling factor $\gamma(t)/\sigma(t)$, along with adders for the dynamic shift and membrane potential accumulation. The dominant hardware cost is the multiplier: $\text{Cost}_{BN} \approx 0.5 N^2$.
    \item \textbf{IS-SNN Neuron:} The normalization operation is removed during inference. Its datapath consists of an accumulator for membrane potential updates and a comparator for spike generation. The arithmetic cost is therefore approximated by the adder cost: $\text{Cost}_{IS} \approx N$.
\end{itemize}

The theoretical reduction in resource consumption can be formalized as:
\begin{equation}
    \text{Theoretical Reduction} \approx \frac{0.5 N^2 - N}{0.5 N^2} = 1 - \frac{2}{N}
\end{equation}
For $N=32$, this estimate gives a theoretical saving of 93.75\%.

The experimental synthesis result (96.4\%) is slightly higher than this theoretical estimate.
This difference can be attributed to two factors:
\begin{enumerate}
    \item \textbf{Control Overhead Elimination:} Dynamic BN variants require additional multiplexers and control logic to handle time-varying parameters ($A(t), B(t)$), consuming extra LUTs beyond the arithmetic units. IS-SNN removes this control overhead.
    \item \textbf{EDA Optimization:} Modern EDA tools (\eg, Vivado) can perform logic packing and optimization on pure adder chains, used in IS-SNN, more effectively than on complex multiplier-adder structures.
\end{enumerate}

This analysis shows that the reported efficiency gains are consistent with the underlying hardware principles and support the advantage of IS-SNN for hardware-oriented deployment.

\end{document}